\documentclass[nohyperref]{article}

\usepackage{microtype}
\usepackage{graphicx}
\usepackage{subfigure}
\usepackage{booktabs} 
\usepackage{url}

\usepackage{listings}
\usepackage[thinc]{esdiff}
\usepackage{hyperref}
\usepackage{tikz}
\usetikzlibrary{bayesnet}
\usetikzlibrary{arrows}
\usetikzlibrary{positioning}
\usetikzlibrary{arrows,automata} 
\usetikzlibrary{external}
\usetikzlibrary{arrows,positioning} 
\tikzset{
    >=stealth',
    punkt/.style={
          rectangle,
          rounded corners,
          draw=black, very thick,
          text width=6.5em,
          minimum height=2em,
          text centered},
    pil/.style={
          ->,
          thick,
          shorten <=2pt,
          shorten >=2pt,}
}

\definecolor{codegreen}{rgb}{0,0.6,0}
\definecolor{codegray}{rgb}{0.5,0.5,0.5}
\definecolor{codepurple}{rgb}{0.58,0,0.82}
\definecolor{backcolour}{rgb}{0.95,0.95,0.92}

\lstdefinestyle{mystyle}{
    backgroundcolor=\color{backcolour},   
    commentstyle=\color{codegreen},
    keywordstyle=\color{magenta},
    numberstyle=\tiny\color{codegray},
    stringstyle=\color{codepurple},
    basicstyle=\ttfamily\footnotesize,
    breakatwhitespace=false,         
    breaklines=true,                 
    captionpos=b,                    
    keepspaces=true,                 
    numbers=left,                    
    numbersep=5pt,                  
    showspaces=false,                
    showstringspaces=false,
    showtabs=false,                  
    tabsize=2
}
\usepackage[accepted]{icml2022}

\usepackage{amsmath,amsfonts,bm}

\newcommand{\D}{\mathcal{D}}

\newcommand{\R}{\mathbb{R}}

\newcommand{\E}{\mathbb{E}}

\newcommand{\calX}{\mathcal{X}}

\newcommand{\calY}{\mathcal{Y}}
\newcommand{\calL}{\mathcal{L}}

\newcommand{\calZ}{\mathcal{Z}}

\newcommand{\RN}[1]{%
  \textup{\uppercase\expandafter{\romannumeral#1}}%
}

\usepackage{amsmath}
\usepackage{amssymb}
\usepackage{mathtools}
\usepackage{amsthm}

\usepackage[capitalize,noabbrev]{cleveref}

\theoremstyle{plain}
\newtheorem{theorem}{Theorem}[section]
\newtheorem{proposition}[theorem]{Proposition}

\theoremstyle{definition}

\theoremstyle{remark}

\usepackage[textsize=tiny]{todonotes}

\icmltitlerunning{Fair Representation Learning through Implicit Path Alignment}

\begin{document}

\twocolumn[
\icmltitle{Fair Representation Learning through Implicit Path Alignment}



\icmlsetsymbol{equal}{*}

\begin{icmlauthorlist}
\icmlauthor{Changjian Shui}{yyy}
\icmlauthor{Qi Chen}{yyy}
\icmlauthor{Jiaqi Li}{sch}
\icmlauthor{Boyu Wang}{sch}
\icmlauthor{Christian Gagné}{yyy,comp}
\end{icmlauthorlist}

\icmlaffiliation{yyy}{Universit\'e Laval, Qu\'ebec, Canada}
\icmlaffiliation{sch}{University of Western Ontario, Ontario, Canada}
\icmlaffiliation{comp}{Canada CIFAR AI Chair, Mila}

\icmlcorrespondingauthor{Boyu Wang}{bwang@csd.uwo.ca}
\icmlcorrespondingauthor{Christian Gagn\'e}{christian.gagne@gel.ulaval.ca}

\icmlkeywords{Machine Learning, ICML}

\vskip 0.3in
]



\printAffiliationsAndNotice{}  

\begin{abstract}
We consider a fair representation learning perspective, where optimal predictors, on top of the data representation, are ensured to be invariant with respect to different sub-groups. Specifically, we formulate this intuition as a bi-level optimization, where the representation is learned in the outer-loop, and invariant optimal group predictors are updated in the inner-loop. Moreover, the proposed bi-level objective is demonstrated to fulfill the \emph{sufficiency rule}, which is desirable in various practical scenarios but was not commonly studied in the fair learning. Besides, to avoid the high computational and memory cost of differentiating in the inner-loop of bi-level objective, we propose an implicit path alignment algorithm, which only relies on the solution of inner optimization and the implicit differentiation rather than the exact optimization path. We further analyze the error gap of the implicit approach and empirically validate the proposed method in both classification and regression settings. Experimental results show the consistently better trade-off in prediction performance and fairness measurement.
\end{abstract}

\section{Introduction}
Machine learning has been widely used in the real world decision-making practice such as job candidate screening \citep{raghavan2020mitigating}.
However, it has been observed that learning algorithms treated some groups of population unfavorably, for example, predicting the likelihood of crime on the grounds of ethnicity, gender or age \citep{hardt2016equality}. 
To that end, algorithmic fairness, which aims to mitigate the prediction bias for the \emph{protected feature} such as gender, has recently received tremendous attentions.  

With the advancements of deep learning \citep{lecun2015deep}, fair representation learning \citep{zemel2013learning} has been recently highlighted. Specifically, the learned fair representation can easily transfer the unbiased prior knowledge to various downstream learning-tasks. For example, in language understanding, the fair embedding provides both useful and unbiased representation for different goals such as translation or recommendation \citep{chang-etal-2019-bias,ethayarajh-2020-classifier}. Besides, it has been investigated in other scenarios such as computer vision \citep{kehrenberg2020null} and intelligent health \citep{10.3389/frai.2020.561802}.  

Typically, fair representation learning is realized by introducing fair constraints during the training. Consequently, a number of
fair notions for various goals have been proposed.
Specifically, most existing approaches in classification or regression use \emph{independence} or \emph{separation} rule (see Sec.\ref{sec:sec2} and references therein) \citep{madras2018learning,song2019learning, chzhen2020fair}.
However, in a variety of applications, independence or separation are not always appropriate, and other fair notions such as \emph{sufficiency rule} \cite{chouldechova2017fair} are preferred.  Intuitively, given the output of the algorithm $\hat{Y}$, the \emph{sufficiency rule} ensures the conditional expectation of label $\E[Y|\hat{Y}]$ is invariant across the different sub-groups (see Sec.\ref{sec:sec2} for the formal definition). 

\begin{figure*}[t]  
\centering  
\subfigure[Unfair Representation]  
{\begin{tikzpicture}[x=0.8cm,y=0.3cm]
\node (h_0) at (0,0){\textbullet};
\node (h_0p) at (-0.4,-0.4){$h^{(0)}$};
\node (h_1) [blue] at (4, -1){\textbullet};
\node (h_1p) [blue] at (4.5,-1.2){$h^{\star}_0$};
\node (h_2) [red] at (1.5, 1){\textbullet};
\node (h_2p) [red] at (2, 1.5){$h^{\star}_1$};
\node (h_3) at (2.75,-0.1){\textbullet};
\node (h_3p) at (3.2,0.2){$h$};
\path (h_0) edge[pil, dashed, color=blue, bend right=20]node[below right, pos=0.35]{$-\nabla_{h}\mathcal{L}_0(h,\lambda)$}(h_1);
\path (h_0) edge[pil, dashed, color=red, bend left=20]node[auto]{$-\nabla_{h}\mathcal{L}_0(h,\lambda)$}(h_2);
\edge[pil]{h_0}{h_3};
\end{tikzpicture}}
\subfigure[Explicit Path]  
{\begin{tikzpicture}[x=0.9cm,y=0.3cm]
\node (h_0) at (0,0){\textbullet};
\node (h_0p) at (-0.4,-0.4){$h^{(0)}$};
\node (h_1) [blue] at (4, -0.2){\textbullet};
\node (h_2) [red] at (4, 1.2){\textbullet};
\node (h_1p) [blue] at (4.5,-0.2){$h^{\star}_0$};
\node (h_2p) [red] at (4.5, 1.6){$h^{\star}_1$};
\path (h_0) edge[pil, dashed, color=blue, bend right=20]node[below right, pos=0.35]{$-\sum_{t} \nabla_{h}\mathcal{L}_0(h^{(t)},\lambda)$}(h_1);
\path (h_0) edge[pil, dashed, color=red, bend right=15,pos=0.42]node[auto]{$-\sum_{t} \nabla_{h}\mathcal{L}_1(h^{(t)},\lambda)$}(h_2);
\end{tikzpicture}}
\subfigure[Implicit Path]{
\begin{tikzpicture}[x=0.9cm,y=0.3cm]
\node (h_0) at (0,0){\textbullet};
\node (h_0p) at (-0.4,-0.4){$h^{(0)}$};
\node (h_1) [blue] at (4, -0.2){\textbullet};
\node (h_2) [red] at (4, 1.2){\textbullet};
\node (h_1p) [blue] at (4.5,-0.5){$h^{\star}_0$};
\node (h_2p) [red] at (4.5, 2.0){$h^{\star}_1$};
\path (h_0) edge[pil, dashed, color=blue, bend right=20](h_1);
\path (h_1) edge[pil,color=orange, dashed](h_0);
\path (h_0) edge[pil, dashed, color=red, bend right=20](h_2);
\path (h_2) edge[pil,color=orange, dashed](h_0);
\end{tikzpicture}}
\caption{Illustration of explicit and implicit path. (a)~Unfair representation leads to different optimization paths and non-invariant optimal predictors on the latent space $\calZ$. (b)~The fair representation learning ensures the invariant optimal predictor w.r.t. different sub-groups on $\calZ$ (encouraging $h^{\star}_0 = h^{\star}_1$). Since the gradient based approach is adopted to optimize $h$, the explicit path alignment aims to learn a representation $\lambda(x)$ to enforce the identical optimization path (i.e, identical \textcolor{blue}{blue} and \textcolor{red}{red} curve) w.r.t. $h$. (c)~The proposed implicit path alignment only requires the last iteration point and approximate the gradient w.r.t. $\lambda$ from the last update of $h$ (\textcolor{orange}{orange} arrow).}\label{fig:2}
\end{figure*}
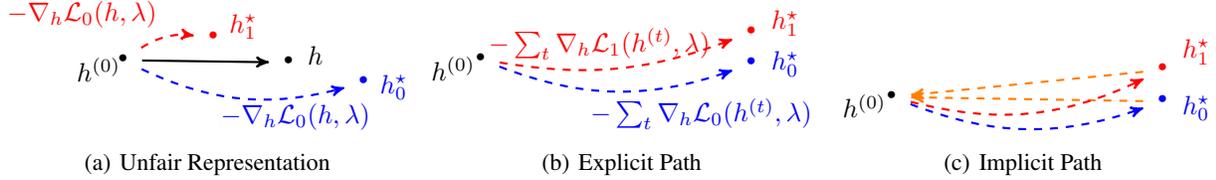

In practice, the negligence of sufficiency rule can lead to the significant bias in intelligent health. For example, health systems rely on commercial algorithms to identify and assist patients with complex health needs. Such algorithms output a score of healthcare needs, where a higher score indicates that the patient is sicker and requires additional care. Notably, \citet{doi:10.1126/science.aax2342} reveals a industry-wide used algorithm that affects millions of patients, exhibits significant racial bias. Under the same predicted score $\hat{Y}=t$, Black patients are considerably sicker than White patients ($\E_{\text{black}}[Y|\hat{Y}=t] > \E_{\text{white}}[Y|\hat{Y}=t]$).  \citet{doi:10.1126/science.aax2342} further points out that eliminating this disparity would increase the percentage of Black patients receiving additional healthcare from $17.7\%$ to $46.5\%$.

From  the algorithmic perspective, sufficiency rule is generally non-compatible to independence or separation, as demonstrated in Sec.\ref{sec:sec2}, indicating that existing fair algorithms for independence or separation do not improve or even worsen the sufficiency rule.    
Therefore fair representation learning w.r.t. the sufficiency rule is important and promising in both practice and algorithmic development. 



In this paper, we propose a framework to address the \emph{sufficiency rule} via the following principle: given a fixed representation function, if the \emph{optimal} predictor that learned on the embedding space are \emph{invariant} to different sub-groups, then the corresponding representation function is fair. The principle is further illustrated in Fig.~\ref{fig:2}(a): when the representation function $\lambda:\calX\to\calZ$ is unfair and we adopt gradient descent to learn the predictor $h:\calZ\to R$. The optimal predictors of different sub-groups (\textcolor{blue}{blue}, \textcolor{red}{red}) are not invariant, yielding biased predictions. Intuitively, the optimal predictor for each subgroup approximates the conditional expectation, which encourages the sufficiency. We will justify such an principle ensures that learned representation could satisfy the sufficiency rule under proper assumptions, showing in Proposition \ref{props1}.

The aforementioned principle can be naturally formulated as a bi-level optimization problem, where we aim to adjust the representation $\lambda$ (in the outer-loop) to satisfy the invariant optimal predictor $h$ (in the inner-loop). Based on this, when we adopt the gradient-based approach in solving the bi-level objective, a straightforward solution is to learn the representation $\lambda$ to fulfill the identical \emph{explicit} gradient-descent directions in learning optimal predictor $h^{\star}$ of different groups, shown in Fig.~\ref{fig:2}(b). Clearly, if the inner gradient descent step of each sub-group is identical, their final predictors (as the approximation of $h^{\star}$) will be surely invariant. 
However, the corresponding algorithmic realization is challenging in deep learning: 1) It requires storing the whole gradient steps, which induces a high memory burden. 2) the embedding function $\lambda$ is optimized via backpropagation from the whole gradient optimization path, which induces a high computational complexity.

To address this, we propose an implicit path alignment, shown in Fig.~\ref{fig:2}(c). Namely, we only consider the final ($t$-th) update of the predictor $h^{(t)}$, then we update representation function $\lambda$ by approximating its gradient at point $h^{(t)}$ through the implicit function \citep{bengio2000gradient}. By using the gradient approximation, we do not need to store the whole gradient steps and conduct the backpropagation through the entire optimization path. Overall, contributions in this paper are as follows:

\textbf{Fair representation learning for the sufficiency rule} The proposed fair-representation approach is proved to satisfy the sufficiency rule in both classification and regression. We also find such a criteria is intrinsically consistent with the recent proposed Invariant Risk Minimization  \citep{arjovsky2019invariant,buhlmann2020invariance}, 
which aims to preserve the invariant correlations between the embedding (or representation) and true label. Intuitively, if such correlations are robust and not influenced by the specific sub-group, the learned representation is somehow fair.
    
\textbf{Efficient algorithm}  We propose an implicit path alignment algorithm to learn the fair representation, which address the prohibitive memory and computational cost in the original bi-level objective. We analyze the approximation error gap of the proposed implicit algorithm, which induces a trade-off between the correct gradient estimation and fairness.

\textbf{Improved fairness in classification and regression} We evaluate the implicit algorithm in classification and regression with tabular, computer vision and NLP datasets, where the implicit algorithm effectively improves the fairness.

\section{Sufficiency rule}\label{sec:sec2}
We denote $X\in\calX$ as the input, $Y\in\calY$ as the ground truth label, and $\hat{Y}\in\calY$ as algorithm's output. Following the previous work in fair representation learning \cite{madras2018learning}, we consider binary protected feature or two sub-groups with corresponding distributions $\D_0$ and $\D_1$. Then according to \citep{pmlr-v97-liu19f}, the sufficiency rule is defined as:
\begin{equation}
     \E_{\D_0}[Y|\hat{Y}=t] = \E_{\D_1}[Y|\hat{Y}=t],~~\forall t\in\calY 
     \label{eq:sufficient_rule}
\end{equation}
Eq.(\ref{eq:sufficient_rule}) shows that the conditional expectation of ground truth label $Y$ are identical for $\D_0$, $\D_1$, given the same prediction output $t$. Based on Eq.(\ref{eq:sufficient_rule}), we propose the sufficiency gap as the metric to measure the fairness. Since we aim to evaluate this in both binary classification ($Y\in\{-1,1\}$) and regression ($Y\in\R$), the sufficiency gaps are separately defined.  

\paragraph{Sufficiency gap in binary classification}  Based on the sufficiency rule, the sufficiency gap in binary classification is naturally defined as:
\begin{small}
\begin{equation*}
    \Delta \text{Suf}_{C} = \frac{1}{2}\sum_{y\in\{-1,1\}} |\D_0 (Y=y|\hat{Y}=y)-\D_1 (Y=y|\hat{Y}=y)|
    \label{eq:sufficient_class}
\end{equation*}
\end{small}

$\Delta\text{Suf}_{C}\in[0,1]$ encourages two sub-groups with identical Positive predicted value (PPV) and Negative predicted value (NPV). To better understand this metric, consider the example of healthcare system, which outputs only binary score: \emph{High Risk} or \emph{Low Risk}.  \citet{doi:10.1126/science.aax2342} essentially revealed $\D_{\text{black}}(Y=\text{High~Risk}|\hat{Y}=\text{Low~Risk}) > \D_{\text{white}}(Y=\text{High~Risk}|\hat{Y}=\text{Low~Risk})$: the severity of illness in Black patients is actually underestimated. Thus if $\Delta \text{Suf}_{C}$ is small, the racial discrimination will be remedied.

\paragraph{Sufficiency gap in regression} Based on  \citep{kuleshov2018accurate}, the sufficiency gap in regression is defined as:
\begin{equation*}
     \Delta \text{Suf}_{R} =  \int_{t\in\calY} |\D_0 (Y\leq t|\hat{Y} \leq t) - \D_1 (Y \leq t |\hat{Y} \leq t)| dt
     \label{eq:sufficient_regression}
\end{equation*}

\begin{figure}[t]
\centering
\includegraphics[width=0.3\textwidth]{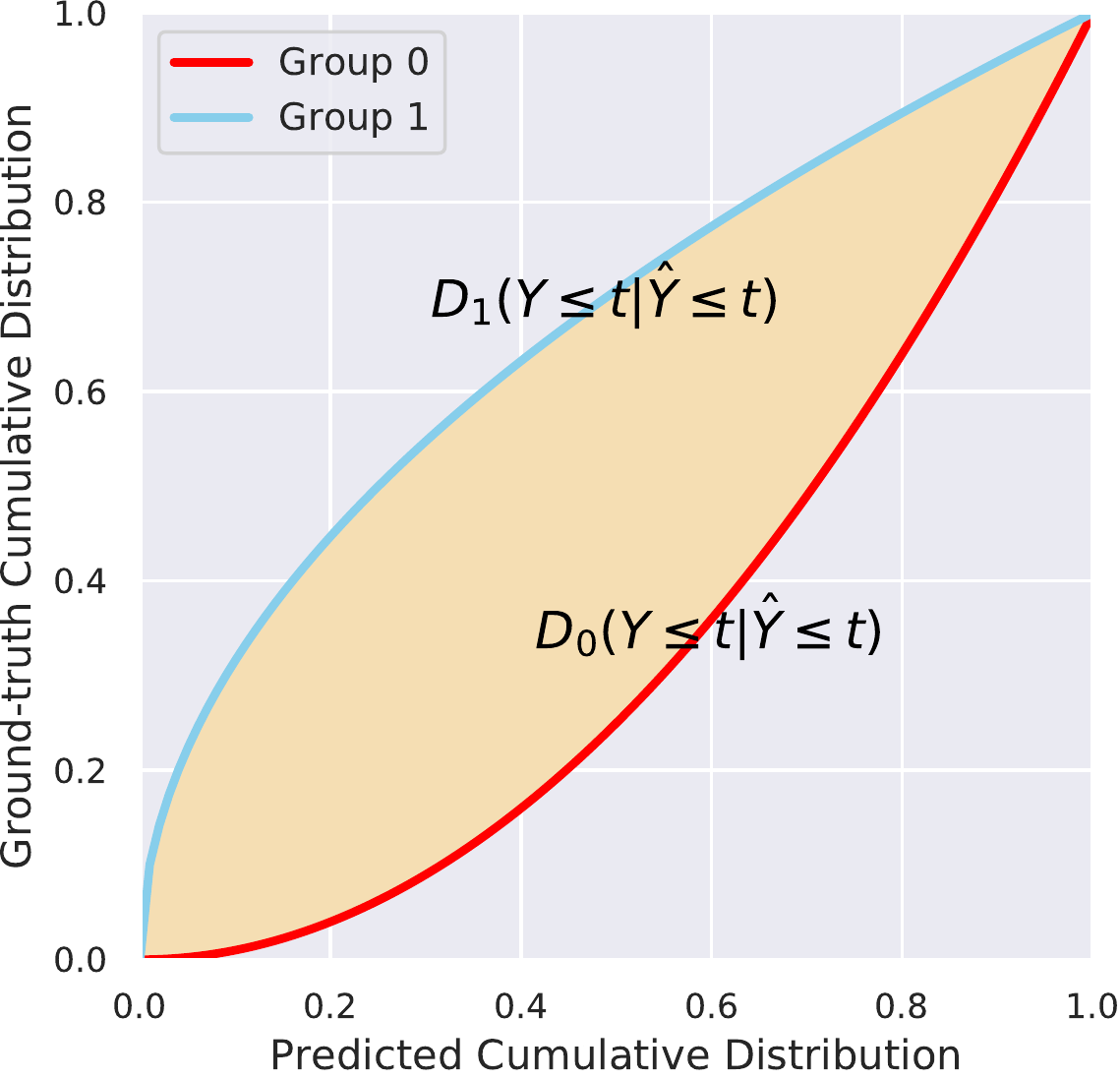}
\caption{Illustrative example of Sufficiency gap ($\Delta\text{Suf}_{R}$) in regression}
\label{fig:pp_in_regression}
\end{figure}

An illustrative example depicts in Fig.~\ref{fig:pp_in_regression}. Specifically, $\Delta\text{Suf}_{R}\in[0,1]$ is an approximation of $|\D_0 (Y = y|\hat{Y} = y)-\D_1(Y = y|\hat{Y} = y)|,~\forall y\in\R$, since the latter is difficult to estimate since $Y$ is continuous. We also adopt the healthcare example to understand this metric: assuming the health system outputs a real-value healthcare score $\hat{Y}=t$ (higher indicates sicker), \citet{doi:10.1126/science.aax2342,sjoding2020racial} observed $\D_{\text{black}} (Y >  t|\hat{Y} \leq t) > \D_{\text{white}} (Y > t |\hat{Y} \leq t)$. Namely, for all the patients with the predicted healthcare score lower than $t$, the actual sicker proportion ($Y>t$) in Black patients is significantly higher than White patients. Therefore a small $\Delta\text{Suf}_{R}$ suggests an improved disparity w.r.t. the sufficient rule.

\paragraph{Relation to other fair rules} We briefly compare the Sufficiency rule with widely adopted Independence and Separation rule in binary classification. The detailed justifications and comparisons are shown in Appendix.

\textbf{Independence rule} is defined as:
\begin{equation*}
    \E_{\D_0}[\hat{Y}] = \E_{\D_1}[\hat{Y}], 
\end{equation*}
In binary classification, the Independence rule is also referred as \emph{demographic parity} (DP) \cite{zemel2013learning}. We can further justify that if $\D_0(Y=y)\neq \D_1(Y=y)$ (i.e, different label distribution in the sub-groups), the Sufficiency and Independence rule cannot both hold. 

\textbf{Separation Rule} is defined as:
\begin{equation*}
    \E_{\D_0}[\hat{Y}|Y=t] = \E_{\D_1}[\hat{Y}|Y=t], ~~\forall t\in\calY
\end{equation*}

In binary classification, the Separation rule is also denoted as \emph{Equalized Odds} (EO) \cite{hardt2016equality}. \citet{barocas-hardt-narayanan} further justified that if $\D_0(Y=y)\neq \D_1(Y=y)$ and the joint distribution of $(Y,\hat{Y})$ has positive probability in $\D_0,\D_1$, the Sufficiency and Separation rule cannot both hold.

\section{Fair representation learning as a bi-level optimization}
We denote the representation function $\lambda$ that maps the input $X$ into the latent variable $Z\in\calZ$, the prediction function $h$ such that $h:\calZ\to\R$ for regression and $h:\calZ\to \{-1,1\}$ for binary classification. We denote the prediction loss as $\ell$, the prediction loss on sub-group $\D_0,\D_1$ is expressed as:
\begin{align*}
    & \calL_0(h,\lambda) = \E_{(x,y)\sim\D_0}\ell(h\circ \lambda (x),y) \\
    & \calL_1(h,\lambda) = \E_{(x,y)\sim\D_1}\ell(h\circ \lambda (x),y)
\end{align*}
According to the intuition, we aim to solve the following bi-level objective:
\begin{align*}
    & \min_{\lambda}~\calL_0(h_0^{\star},\lambda) +  \calL_1(h_1^{\star},\lambda) \tag*{(Outer-Loop)}\\
    & \text{s.t.}~~ h_{0}^{\star} = h_{1}^{\star},\tag*{(Inner-Loop)}\\
    & h_{0}^{\star} \in \underset{h}{\text{argmin}}  ~\calL_0(h,\lambda), ~ h_{1}^{\star} \in \underset{h}{\text{argmin}}  ~\calL_1(h,\lambda). 
\end{align*}
In the outer-loop, we aim to find a representation function $\lambda$ for minimizing the prediction error, given the optimal predictor ($h_0^{\star}, h_1^{\star}$) on the embedding space $\calZ$. As for the inner-loop, given a fixed representation $\lambda$, $h_{0}^{\star}$, $h_{1}^{\star}$ are the optimal predictor for each sub-group. The constraints $h_0^{\star}=h_1^{\star}$ additionally encourage the invariant optimal predictors from $\D_0$, $\D_1$. 

\paragraph{Relation to explicit path alignment} In deep learning, we adopt gradient-based approaches to minimize the loss, therefore $h^{\star}$ in the inner-loop is approximated as $h^{(t+1)}$, the $t$-th update in the gradient descent:
$h_0^{\star}\approx h^{(0)} - \sum_{t} \nabla_{h} \calL_0 (h^{(t)},\lambda)$, $h_1^{\star}\approx h^{(0)} - \sum_{t} \nabla_{h} \calL_1 (h^{(t)},\lambda)$, where $h^{(0)}$ is the common initialization. Thus the invariant optimal predictor is equivalent to:
\begin{equation*}
    \sum_{t} \nabla_{h} \calL_0 (h^{(t)},\lambda) = \sum_{t} \nabla_{h} \calL_1 (h^{(t)},\lambda).
\end{equation*}
The aforementioned equation suggests learning a representation $\lambda$ that ensures the identical optimization path w.r.t. $h$ for each sub-group, which recovers the explicit path alignment. 

\textbf{Relation to Sufficiency rule} We further demonstrate the relation between the bi-level objective and Sufficiency rule.
\begin{proposition}\label{props1}
If we specify the prediction loss $\ell$ as logistic regression loss in the classification $\log(1+\exp(-yh(z)))$ with $\calY=\{-1,1\}$ and 
the square loss in the regression $(h(z)-y)^2$ with $\calY\subset\R$. Then minimizing the inner-loop loss is equivalent to:
\begin{align*}
    & \E_{\D_0}[Y|Z=z] = \E_{\D_1}[Y|Z=z], \\
    & \E_{\D_0}[Y|\hat{Y} = h^{\star}(z)] = \E_{\D_1}[Y|\hat{Y} = h^{\star}(z)]
\end{align*}
where $h^{\star}=h_0^{\star}=h_1^{\star}$ and $z=\lambda(x)$.
\end{proposition}
Proposition \ref{props1} demonstrates that the objective of inner-loop loss fulfills the sufficiency rule in both binary classification and regression. 

\section{Proposed Algorithms}
We propose the implicit alignment in deep learning, where $\lambda$ and $h$ are implemented by the neural network. We also reformulate as the original objective through \emph{Lagrangian relaxation}:
\begin{align}
    & \min_{\lambda}~\calL_0(h_0^{\star},\lambda) +  \calL_1(h_1^{\star},\lambda) +\frac{\kappa}{2} \|h_{0}^{\star} - h_{1}^{\star}\|^2_2 \tag*{(Outer-Loop)}\\
    & \text{s.t.}~~h_{0}^{\star} \in \underset{h}{\text{argmin}}  ~\calL_0(h,\lambda), ~ h_{1}^{\star} \in \underset{h}{\text{argmin}}  ~\calL_1(h,\lambda), \tag*{(Inner-Loop)}
\end{align}
where the introduced $\kappa>0$ is the coefficient to control the fairness, with a sufficient large $\kappa$ ensuring $h_0^{\star}\approx h_1^{\star}$. Then we drive the approximated gradient w.r.t. $\lambda$, which contains the following key elements.

\textbf{Solving the inner optimization} Given a fixed representation $\lambda$, we find $h_0^{\epsilon}$, $h_1^{\epsilon}$ such that:
\[\|h_0^{\star}-h_0^{\epsilon}\|\leq \epsilon, ~~~~ \|h_1^{\star}-h_1^{\epsilon}\| \leq \epsilon,
\]
where $\epsilon$ is the optimization tolerance. Besides, $h_1^{\star}$ and $h_1^{\epsilon}$ are essentially the function of $\lambda$, i.e., $h_1^{\epsilon}$ depends on the predefined representation function $\lambda$. It is worth mentioning that the optimization tolerance $\epsilon$ is realistic. E.g, consider a fixed representation and one-layer predictor $h_0,h_1$, the optimization will be convex.

\textbf{Computing the gradient of $\lambda$} Given the approximate solution $h_0^{\epsilon}$, $h_1^{\epsilon}$, we can compute the gradient w.r.t. $\lambda$ (referred as $\tilde{\text{grad}}(\lambda)$) \footnote{We denote the ground truth gradient as $\text{grad}(\lambda)$ if we adopt optimal predictor $h_0^{\star}, h_1^{\star}$ in the computation.} in the outer-loop:
\begin{align*}
 \tilde{\text{grad}}(\lambda) 
=  &  \nabla_{\lambda}\calL_0(h_0^{\epsilon},\lambda) +  \nabla_{\lambda}\calL_1(h_1^{\epsilon},\lambda) \\
& \left(\nabla_{\lambda} h^{\epsilon}_0\right)^T \left(\nabla_{h_0}\calL_0(h_0^{\epsilon},\lambda) 
+  \kappa(h^{\epsilon}_0-h^{\epsilon}_1)\right) \\
&  + \left(\nabla_{\lambda} h^{\epsilon}_1\right)^T \left(\nabla_{h_1}\calL_1(h_1^{\epsilon},\lambda)-\kappa (h^{\epsilon}_0-h^{\epsilon}_1)\right).
\end{align*}
Where $\nabla_{h_0}\calL_0(h_0^{\epsilon},\lambda)$ is the partial derivative in the loss w.r.t. the first term (about $h_0$), evaluated at $h_0^{\epsilon}$. Also $\nabla_{\lambda}\calL_0(h_0^{\epsilon},\lambda)$ is the partial derivative w.r.t. the second term (about $\lambda$).

\textbf{Implicit function for approximating the gradient} In order to compute $\tilde{\text{grad}}(\lambda)$ in \verb+autograd+, we need to estimate $\nabla_{\lambda} h^{\epsilon}_0$ and $\nabla_{\lambda} h^{\epsilon}_1$. We herein adopt the implicit function \citep{bengio2000gradient} to approximate $\nabla_{\lambda} h^{\epsilon}_0$, which has been adopted in the hyperparameter optimization \citep{pedregosa2016hyperparameter} and meta-learning \citep{rajeswaran2019meta}. 

Concretely, if the prediction loss is smooth and there exist stationary points to achieve optimal, we have: $\nabla_{h_0} \calL_0(h_0^{\star}(\lambda),\lambda)= 0, \nabla_{h_1} \calL_0(h_1^{\star}(\lambda),\lambda)= 0$. Then differentiating w.r.t. $\lambda$ will induce: $\mathbf{d}\left(\nabla_{h_0} \calL_0(h_0^{\star}(\lambda),\lambda)\right)/\mathbf{d}\lambda=\nabla_{h_0}^2 \calL_0(h_0^{\star},\lambda)\nabla_{\lambda}h_0^{\star} + \nabla_{\lambda} \nabla_{h_0} \calL_0(h_0^{\star},\lambda)=0$.\footnote{$\mathbf{d}(\cdot)/\mathbf{d}\lambda$ denotes the total derivative.} Thus we have $\nabla_{\lambda} h_0^{\star} = -\left(\nabla^2_{h_0} \calL_0(h_0^{*}, \lambda) \right)^{-1} \left(\nabla_{\lambda}\nabla_{h_0}\calL_0(h_0^{*}, \lambda) \right)$, where the Hessian matrix $\nabla^2_{h_0} \calL_0(h_0^{*}, \lambda)$ is assumed to be invertible.

Through the implicit function, we can approximate $\nabla_{\lambda} h^{\epsilon}_0$ as:
\[
\nabla_{\lambda} h_0^{\epsilon} \approx -\left(\nabla^2_{h_0} \calL_0(h_0^{\epsilon},\lambda) \right)^{-1} \left(\nabla_{\lambda}\nabla_{h_0}\calL_0(h_0^{\epsilon},\lambda) \right)
\]
As for $\nabla_{\lambda} h^{\epsilon}_1$, we have the similar result: $\nabla_{\lambda} h_1^{\epsilon} \approx -\left(\nabla^2_{h_1} \calL_1(h_1^{\epsilon},\lambda) \right)^{-1} \left(\nabla_{\lambda}\nabla_{h_1}\calL_1(h_1^{\epsilon},\lambda) \right)$.

\textbf{Efficient and numerical stable gradient estimation} Plugging in the approximations, the gradient w.r.t $\lambda$ is approximated as:

$\tilde{\text{grad}}(\lambda)  \approx \nabla_{\lambda}\calL_0(h_0^{\epsilon},\lambda) -\left(\nabla_{\lambda}\nabla_{h_0}\calL_0(h_0^{\epsilon},\lambda) \right)^{T}\mathbf{p}_0 
 + \nabla_{\lambda}\calL_1(h_1^{\epsilon},\lambda) - \left(\nabla_{\lambda}\nabla_{h_1}\calL_1(h_1^{\epsilon},\lambda) \right)^{T}\mathbf{p}_1$

Where $\mathbf{p}_0$, $\mathbf{p}_1$ are denoted as the inverse-Hessian vector product with:
\begin{align*}
    & \mathbf{p}_0 = \left(\nabla^2_{h_0} \calL_0(h_0^{\epsilon},\lambda) \right)^{-1} \left(\nabla_{h_0}\calL_0(h_0^{\epsilon},\lambda) + \kappa(h^{\epsilon}_0-h^{\epsilon}_1)\right) \\
    & \mathbf{p}_1 = \left(\nabla^2_{h_1} \calL_1(h_1^{\epsilon},\lambda) \right)^{-1} \left(\nabla_{h_1}\calL_1(h_1^{\epsilon},\lambda)-\kappa (h^{\epsilon}_0-h^{\epsilon}_1)\right).
\end{align*}

However, the current form is still computationally expensive due to the computation of inverse Hessian matrix.  Then computing $\mathbf{p}_0$ and $\mathbf{p}_1$ is equivalent to solve the following quadratic programming (QP):
\begin{align}
    \text{argmin}_{\hat{\textbf{p}}_0} &~~\frac{1}{2} \hat{\textbf{p}}^{T}_0 \left(\nabla^2_{h_0} \calL_0(h_0^{\epsilon},\lambda) \right)\hat{\textbf{p}}_0 \nonumber \\
    & - \hat{\textbf{p}}^{T}_0 \left(\nabla_{h_0}\calL_0(h_0^{\epsilon},\lambda) + \kappa(h^{\epsilon}_0-h^{\epsilon}_1)\right) 
    \label{delta_approx_1}
\end{align}
\begin{align}
    \text{argmin}_{\hat{\textbf{p}}_1} &~~\frac{1}{2} \hat{\textbf{p}}^{T}_1 \left(\nabla^2_{h_1} \calL_1(h_1^{\epsilon},\lambda) \right)\hat{\textbf{p}}_1 \nonumber \\
    & - \hat{\textbf{p}}^{T}_1  \left(\nabla_{h_1}\calL_1(h_1^{\epsilon},\lambda) - \kappa(h^{\epsilon}_0-h^{\epsilon}_1)\right) \label{delta_approx}
\end{align}
Since it is a typical QP problem and we adopt conjugate gradient method \citep{ConjugateGradient1985,rajeswaran2019meta}, which can be updated efficiently through \verb+autograd+ via computing the Hessian-vector product. We additionally suppose the optimization error in the QP as $\delta$, i.e.: $\|\mathbf{p}_0 -\mathbf{p}^{\delta}_0 \|\leq \delta$,  $\|\mathbf{p}_1 -\mathbf{p}^{\delta}_1 \|\leq \delta$, then the gradient w.r.t representation $\lambda$ can be finally expressed as:

$ \tilde{\text{grad}}^{\delta}(\lambda) =  \nabla_{\lambda}\calL_0(h_0^{\epsilon},\lambda) -\left(\nabla_{\lambda}\nabla_{h_0}\calL_0(h_0^{\epsilon},\lambda) \right)^{T}\mathbf{p}^{\delta}_0 + \nabla_{\lambda}\calL_1(h_1^{\epsilon},\lambda) - \left(\nabla_{\lambda}\nabla_{h_1}\calL_1(h_1^{\epsilon},\lambda) \right)^{T}\mathbf{p}^{\delta}_1$

The $\tilde{\text{grad}}^{\delta}(\lambda)$ can be also efficiently estimated through Hessian vector product via \verb+autograd+ without explicitly computing the Hessian matrix.

\paragraph{Proposed algorithm} Based on the key elements, the proposed algorithm is shown in Algo.~\ref{bi_level_main}.
\begin{algorithm}[t]
		\caption{Implicit Path Alignment Algorithm}
		\begin{algorithmic}[1] 
        \STATE {\bfseries Input:} Representation function $\lambda$, predictor $h_0$, $h_1$, datasets from two sub-groups $\D_0$. $\D_1$.
        \FOR{mini-batch of samples from $(\D_0,\D_1)$}
        \STATE Solving the inner-loop optimization with tolerance $\epsilon$. Obtaining $h_0^{\epsilon},h_1^{\epsilon}$.
        \STATE Solving Eq. \eqref{delta_approx_1}, \eqref{delta_approx} with tolerance $\delta$. Obtaining $\mathbf{p}^{\delta}_0$ and $\mathbf{p}^{\delta}_1$.
        \STATE Computing $\tilde{\text{grad}}^{\delta}(\lambda)$ (gradient of representation $\lambda$)
        \STATE Updating $\lambda$ through \verb+autograd+: $\lambda \leftarrow \lambda -\tilde{\text{grad}}^{\delta}(\lambda)$
        \ENDFOR
    \STATE \textbf{Return:}~~$\lambda$, $h_0^{\epsilon}$, $h_1^{\epsilon}$
        \end{algorithmic}
        \label{bi_level_main}
\end{algorithm}
\subsection{The cost of Implicit algorithm: Approximation-Fair Trade-off}
In the proposed objective bi-level loss, a sufficient large $\kappa$ encourages the invariant optimal predictor, yielding the fair results. However, the implicit approach will lead to a biased estimation of the ground truth gradient. We analyze the error gap of the approximation in Theorem \ref{thm:1}.

\begin{theorem}[Approximation Error Gap]\label{thm:1} Suppose that (1) \emph{Smooth Predictive Loss}. The first-order derivatives and second-order derivatives of $\calL$ are Lipschitz continuous; (2) \emph{Non-singular Hessian matrix}. We assume $\nabla_{h_0,h_0}\calL_0(h_0,\lambda), \nabla_{h_1,h_1}\calL_1(h_1,\lambda)$, the Hessian matrix of the inner optimization problem, are invertible. (3) \emph{Bounded representation and predictor function}. We assume the $\lambda$ and $h$ are bounded, i.e., $\|\lambda\|, \|h\|$ are upper bounded by the predefined positive constants. Then the approximation error between the ground truth and algorithmic estimated gradient w.r.t. the representation is be upper bounded by:
\[\|\text{grad}(\lambda)- \tilde{\text{grad}}^{\delta}(\lambda)\| = \mathcal{O}(\kappa\epsilon + \epsilon + \delta).\]
\end{theorem}
The proof is delegated in Appendix \ref{appendix:approx_error}. We also discuss the assumptions to guarantee the convergence of Algorithm 1, shown in Appendix \ref{appendix:glo_convergence}.  

Theorem \ref{thm:1} reveals that the gradient approximation error depends on the two-level optimization tolerance $\epsilon$, $\delta$ and the coefficient of fair constraints $\kappa$. Specifically, the error gap reveals the inherent trade-off in accurate gradient estimation and fair-representation learning. If we fix the optimization tolerance $\epsilon$ and $\delta$, a smaller $\kappa$ indicates a better approximation of the gradient, which yields weak fair constraints. Thus the implicit alignment introduces a trade-off in the prediction performance (i.e., correct approximation of the gradient) and fairness measurement.


\section{Related Work}
\paragraph{Fair Machine Learning} Below we only list the most related work and refer to the survey paper \citep{mehrabi2021survey} for details in the algorithmic fairness. In the \emph{classification}, various methods in learning fair representations have been proposed. Specifically, a common strategy is to introduce the statistical constraints as the regularization during the training, e.g., demographic parity (DP) \citep{zhang2018mitigating,madras2018learning,song2019learning,jiang2020wasserstein, kehrenberg2020null} that encourages the identical output of the representation or equalized odds (EO) \citep{song2019learning,gupta2021controllable} that ensures the identical conditional output of the representation, given the ground truth label $Y$. Another direction is to disentangle the data for factorizing meaningful representations such as \citep{locatello2019fairness,kim2019learning}. Intuitively, the disentangled embedding is independent of the protected feature, thus reflecting a fair representation w.r.t. the independence rule, which can be potentially problematic when the label distributions of sub-groups vary dramatically \citep{zhao2019conditional}.

The concept of fairness has also been extended to the fields beyond classification. For instance, in the \emph{regression} problem \citep{pmlr-v80-komiyama18a, agarwal2019fair}, the bounded group loss has been proposed as the fair measure: if the prediction loss in each sub-group is smaller than $\epsilon$, the regression is $\epsilon$-level fair. In fact, the fair criteria in our paper is \emph{not equivalent} to $\epsilon$-fair. Considering a fixed representation function $\lambda$, the $\epsilon$-level fair does not guarantee the \emph{optimal} and \emph{invariant} predictor for each sub-group and vice versa.

The sufficiency rule has also been discussed in the previous work. Notably,~ \citet{chouldechova2017fair,pmlr-v97-liu19f} proposed the sufficiency gap \emph{in classification} for measuring fairness w.r.t. the sufficiency rule. \cite{pmlr-v97-liu19f} also discussed the relations between the sufficiency gap and probabilistic calibration \citep{guo2017calibration} (referred as calibration gap). According to \citet{pleiss2017fairness}, the calibration rule is a stronger condition than sufficiency rule while it can simultaneously hurt the prediction performance. Throughout this paper, we only consider the sufficiency rule. The triple trade-off between the probabilistic calibration, sufficiency rule and accuracy will be left as future work.

\paragraph{Learning Invariance} The analyzed fair-representation criteria shares a quite similar spirit to the IRM or Invariant Risk Minimization \citep{arjovsky2019invariant,buhlmann2020invariance,creager2021environment}, where an algorithm IRM\_v1 is proposed to enable the out-of-distribution (OOD) generalization. The key difference between our work and \citep{arjovsky2019invariant} lies in the algorithmic aspect: it has been theoretically justified that the originally proposed IRM\_v1 does not necessarily capture the invariance across the environments~\citep{rosenfeld2020risks,shui2022benefits}. By contrast, we aim to solve the bi-level objective in the context of deep-learning and propose an efficient and principled practical algorithm with better empirical performance than IRM\_v1. Besides, based on results of \citep{chen2021iterative}, the proposed algorithm does not provably guarantee the OOD generalization property due to the limited sub-groups ($N=2$) considered within the paper.

\section{Experiments}
\subsection{Experimental setup}
In the paper, we adopt the aforementioned sufficiency gap as fair metrics, where $\hat{Y}$ is denoted as:
\begin{small}
\[
    \hat{Y} = 
\begin{cases}
    h_0^{\epsilon}\circ \lambda(X),& X\in\D_0\\
    h_1^{\epsilon}\circ \lambda(X),& X\in\D_1
\end{cases}
\]
\end{small}
Then in the binary classification, we can estimate $\Delta \text{Suf}_{C}= \frac{1}{2}\sum_{y\in\{-1,+1\}}|\D_0(Y=y|\hat{Y}=y)-\D_1(Y=y|\hat{Y}=y)|$ from the data.

As for regression, the original form $\Delta \text{Suf}_{R} = \int_{t} |\D_0(Y\leq t|\hat{Y}\leq t)-\D_1(Y\leq t|\hat{Y}\leq t)|$ (as shown in Fig.~\ref{fig:pp_in_regression}, Appendix) is difficult to estimate due to the integration term. 
To address this, we sample multiple values $\{t_1,\dots,t_m\}$ and compute its average differences as the approximation of the integration. Namely,
$\Delta \text{Suf}_{R} \approx \frac{1}{m}\sum_{i=1}^{m} |\D_0(Y\leq t_i|\hat{Y} \leq t_i)-\D_1(Y\leq t_i|\hat{Y}\leq t_i)|$. Concretely, for a given $t_i$ in each group, we compute the percentile ($\hat{Y}_0$) at point $t$: $\D_0(\hat{Y}_0\leq t_i)$, then we compute the corresponding ground truth cumulative distribution ($Y$) at the same point $t_i$:  $\D(Y \leq t_i|\hat{Y} \leq t_i)$. Through the aforementioned approximation, we can estimate $|\D_0(Y \leq t_i|\hat{Y} \leq t_i)- \D_1(Y \leq t_i|\hat{Y} \leq t_i)|$. 

\paragraph{Baselines} We consider the baselines that add fairness constraints during the training process. Specifically, we compare our method with (\RN{1}) Empirical Risk Minimization (ERM) that trains the model without considering fairness; (\RN{2}) Adversarial Debiasing (referred as adv\_debias) \citep{zhang2018mitigating}; (\RN{3}) Fair Mix-up  \citep{chuang2021fair}, a recent data-augmentation and effective approach in the fair representation learning. In fact, the baselines (\RN{2}) and (\RN{3}) are based on Independence rule or Demographic-Parity (DP), which is designed to demonstrate the general non-compatibility in addressing the sufficiency rule.

Besides, we include two additional baselines that have the similar objective but different algorithmic realizations.  
(\RN{4}) the original IRM regularization (referred as IRM\_v1) \citep{arjovsky2019invariant}, which adds a gradient penalty to encourage the invariance among the different groups. (\RN{5}) One-step explicit alignment. In the inner-loop optimization, we suppose to conduct a simple one-step gradient descent ($T=1$) for each sub-group, i.e, $h_0^{\star}\approx h_{\text{init}} -\nabla_{h_0} \calL_0 (h_0,\lambda)$, $h_1^{\star}\approx h_{\text{init}} -\nabla_{h_1} \calL_1 (h_1,\lambda)$. Thus in the outer-loop optimization, we add a gradient-incoherence constraint to encourage the identical (one-step) optimization path: $\min_{\lambda} \|\nabla_{h_0} \calL_0 (h_0,\lambda) - \nabla_{h_1} \calL_1 (h_1,\lambda) \|_2^2$. 

All the results are reported by averaging five repetitions and additional experimental details are delegated in the Appendix.

\begin{table*}[t]
\centering
\caption{Fair Classification. Accuracy and $\Delta\text{Suf}_C$ in Toxic comments (left) and CelebA datasets (right)}
\label{tab:fair_class}
\resizebox{0.45\textwidth}{!}{
\begin{tabular}{ccc}\toprule
     \textcolor{blue}{Toxic comments} & Accuracy~($\uparrow$) & $\Delta\text{Suf}_{C}$~($\downarrow$) \\ \midrule
     ERM~(\RN{1}) & 0.768 $\pm$ 0.004 & 0.173 $\pm$ 0.008  \\\midrule
     Adv\_debias~(\RN{2}) & 0.760 $\pm$ 0.008 & 0.291 $\pm$ 0.006 \\
     Mixup~(\RN{3})  & 0.758 $\pm$ 0.003 & 0.343 $\pm$ 0.022 \\ \midrule
     IRM\_v1~(\RN{4}) & 0.753 $\pm$ 0.004 & 0.057 $\pm$ 0.015  \\ 
     One\_step~(\RN{5}) & 0.755 $\pm$ 0.007 & 0.048 $\pm$ 0.008 \\
     Implicit & 0.760 $\pm$ 0.007 & 0.051 $\pm$ 0.012 \\ \bottomrule
     \end{tabular}
}
\resizebox{0.45\textwidth}{!}{
\begin{tabular}{ccc}\toprule
     \textcolor{blue}{CelebA} & Accuracy~($\uparrow$) & $\Delta\text{Suf}_{C}$~($\downarrow$) \\ \midrule
     ERM~(\RN{1}) & 0.780 $\pm$ 0.015 & 0.210 $\pm$ 0.022  \\\midrule
     Adv\_debias~(\RN{2}) & 0.785 $\pm$ 0.022 & 0.165 $\pm$ 0.028 \\
     Mixup~(\RN{3})  & 0.792 $\pm$ 0.011 & 0.160 $\pm$ 0.010 \\ \midrule
     IRM\_v1~(\RN{4}) & 0.795 $\pm$ 0.012 & 0.086 $\pm$ 0.015 \\ 
     One\_step~(\RN{5}) & 0.797 $\pm$ 0.006 & 0.086 $\pm$ 0.012 \\
     Implicit & 0.794 $\pm$ 0.027  & 0.074 $\pm$ 0.020  \\ \bottomrule
     \end{tabular}
}
\end{table*}

\begin{figure*}[h]
\centering
 \centering
  \subfigure[Toxic]{\label{fig:toxic}\includegraphics[width=50mm]{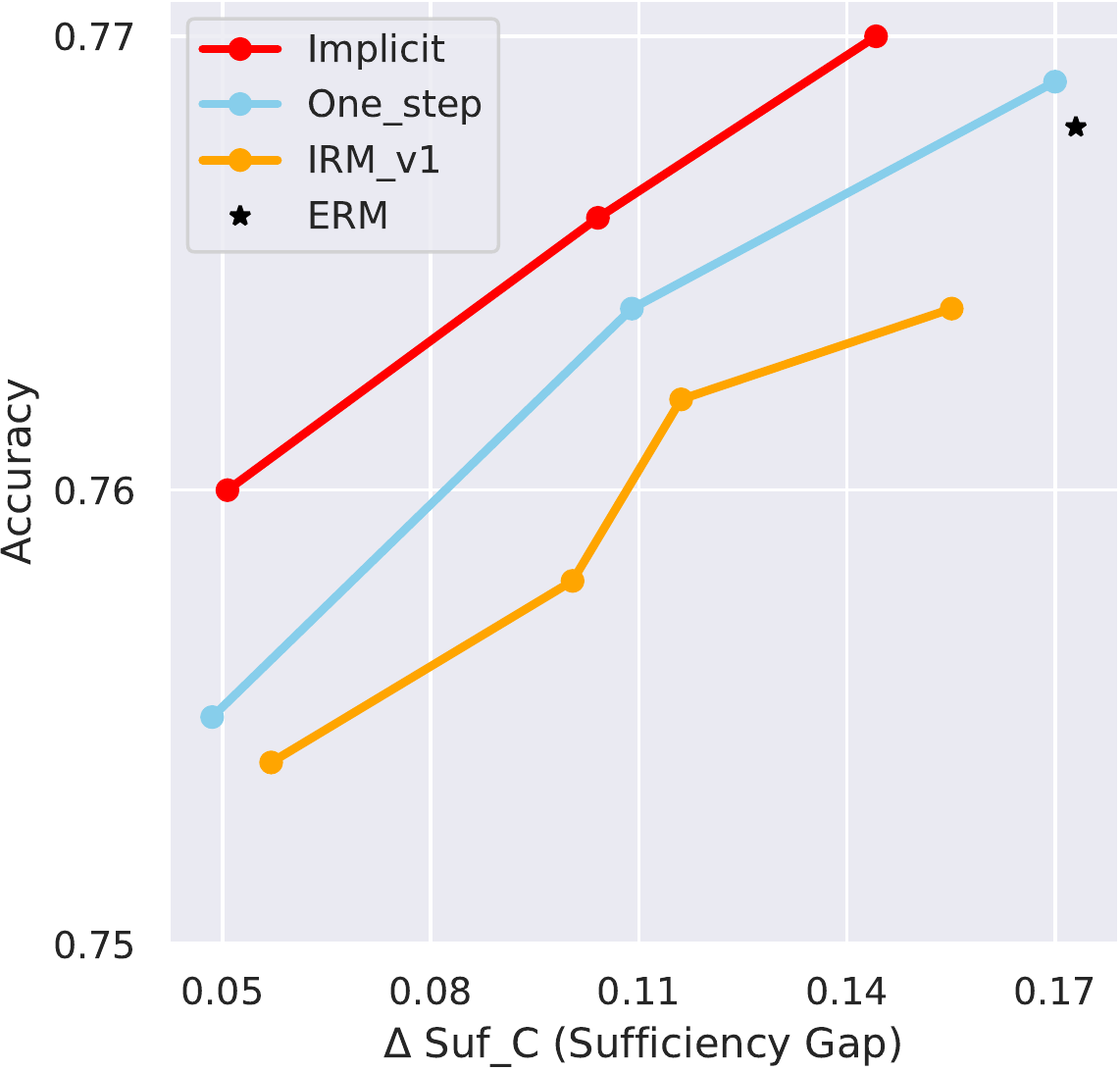}}
  \quad
  \subfigure[CelebA]{\label{fig:celeba}\includegraphics[width=50mm]{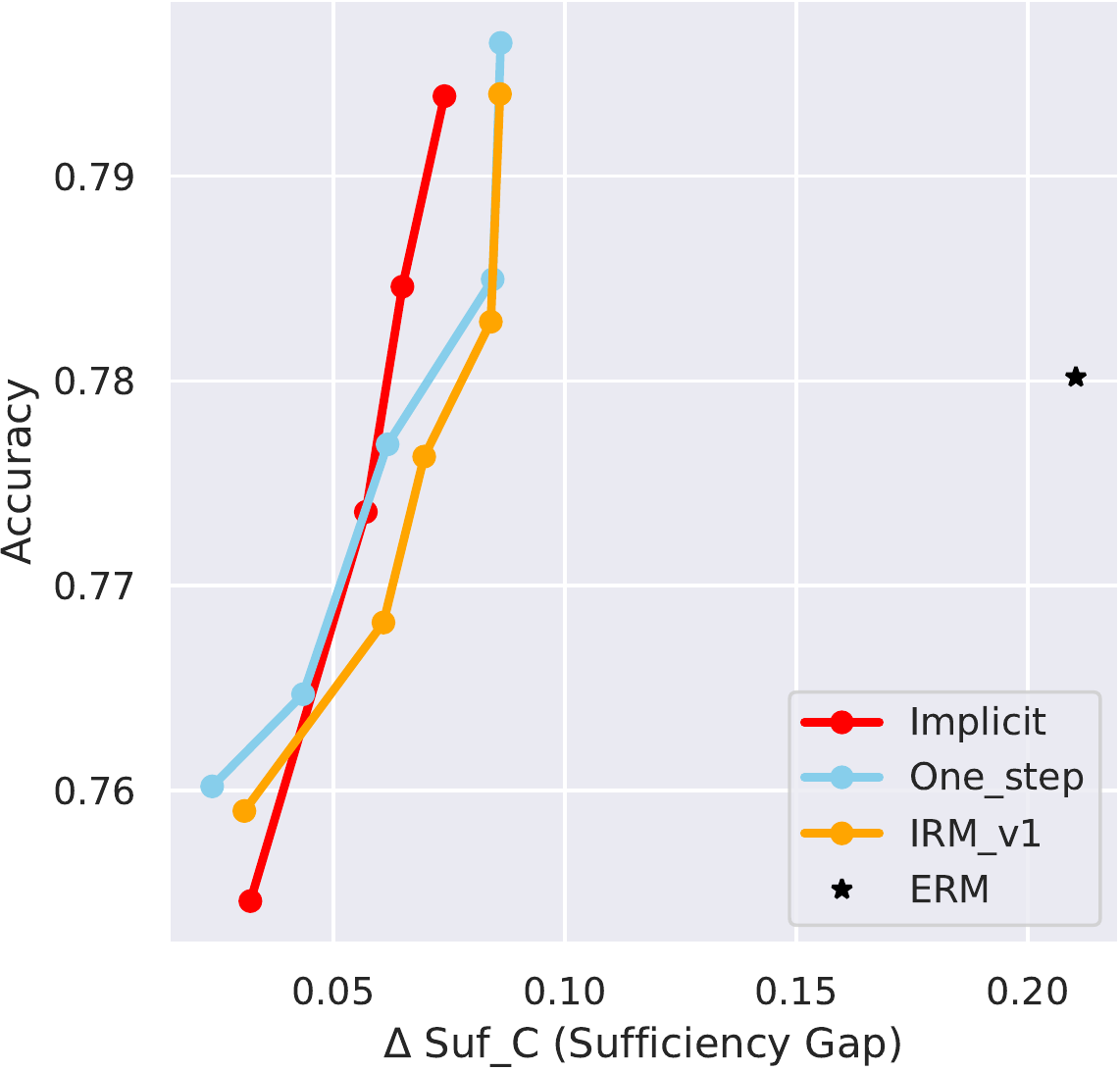}}
  \quad
  \subfigure[Time in CelebA]{\label{fig:ablation_toxic}\includegraphics[width=47mm]{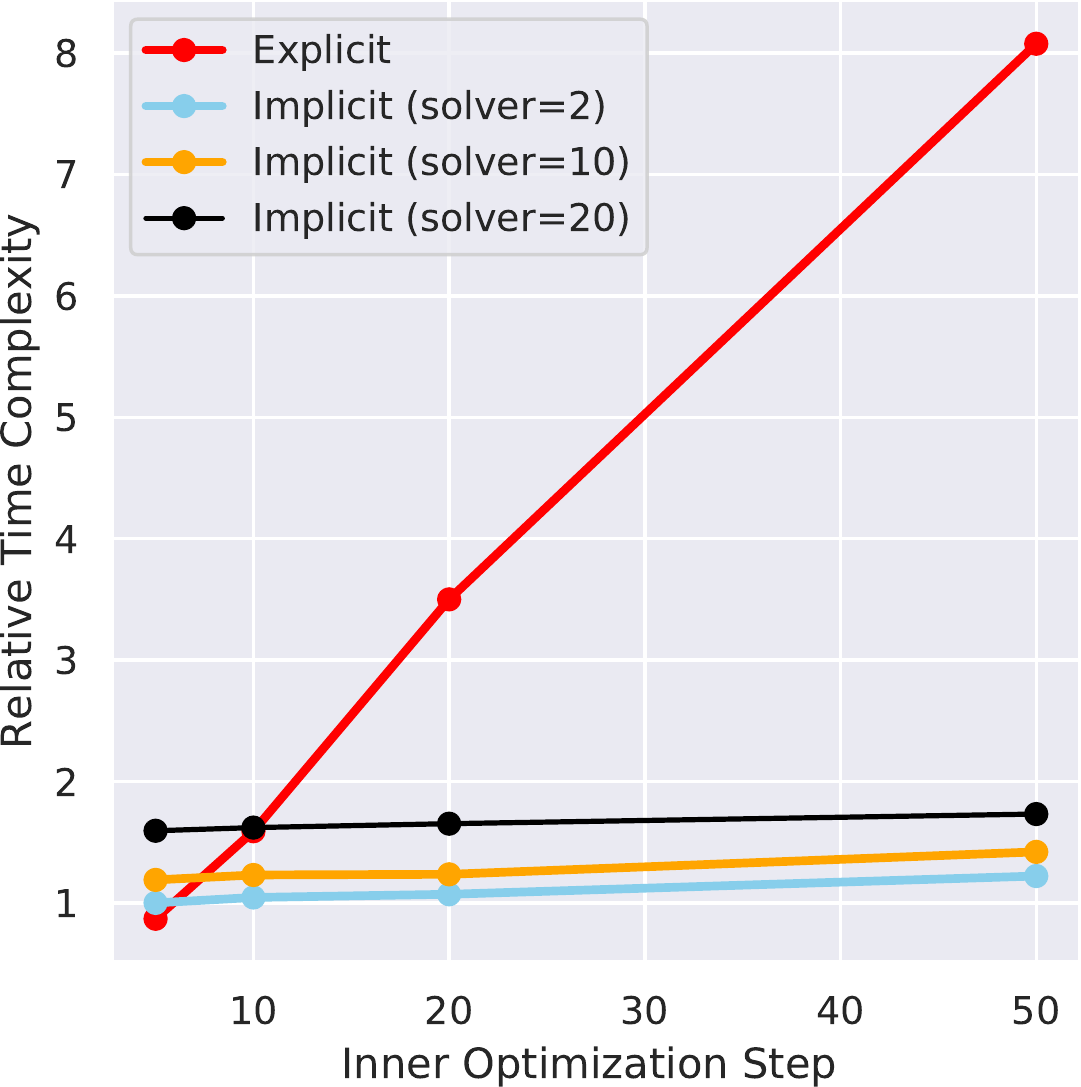}}
  \caption{Fair Classification. (a,b) The Accuracy-Fair trade-off curve in Toxic (a) and CelebA (b) dataset. The implicit approach demonstrated a consistently better trade-off. (c) Running time comparison of Explicit and Implicit alignment in CelebA dataset. Specifically, $\text{solver}=2$ indicates that the conjugate-gradient algorithm is executed 2 iterations. The results show that implicit approach avoids the long back-propagation of the entire inner-optimization path. The time complexity of the explicit approach, on the other hand, increases linearly with the inner-optimization step.}
\end{figure*}

\subsection{Toxic Comments} 
The toxic comments dataset \citep{Jigsaw} is a binary \textbf{classification} task in NLP to predict whether comment is toxic or not. The original label is actually not binary since the comments is decided by multiple annotators, where the labelling discrepancy generally occurs. To this end, we conduct a simple strategy to decide comment is toxic if at least one annotator marks it. In this dataset, a portion of comments have been labeled with identity attributes, including gender and race. It has also been revealed that the race identity (e.g., black) is correlated with the toxicity label, which can lead to the predictive discrimination. Thus we adopted the \emph{race} as the protected feature by selecting two sub-groups of Black and Asian. For the sake of computational simplicity, we first applied the pretrained BERT \citep{devlin2018bert} to extract the word embedding with 748 dimensional vector. Then we adopt representation function $\lambda$ as two fully-connected layers with hidden dimension 200 with Relu activation and classifier $h$ as a linear predictor. We report the test-set sub-group average accuracy and sufficiency gap ($\Delta\text{Suf}_C$) in Tab.~\ref{tab:fair_class} and Fig.~\ref{fig:toxic}.

 

From the results, the Demographic Parity (DP) based fair constraints are non-compatible with the sufficiency rule. Specifically, baseline (\RN{2},\RN{3}) even increase $\Delta\text{Suf}_C$ with higher value than ERM. For the baselines that track the sufficiency rule (\RN{4},\RN{5}), the sufficiency gap $\Delta\text{Suf}_C$ is improved with a similar accuracy, shown in Tab.\ref{tab:fair_class}. We also change the regularization coefficient in (\RN{4},\RN{5}) and $\kappa$ in the implicit approach. We observe that the implicit approach demonstrates a consistent better Accuracy-Fair trade-off, shown in Fig.~\ref{fig:toxic}. 

\subsection{CelebA Dataset} 
The CelebA dataset \citep{liu2015faceattributes} contains around 200K images of celebrity faces, where each image is associated with 40 human-annotated binary attributes including gender, hair color, young, etc. In this paper, we designate \emph{gender} as the protected feature, and \emph{attractive} as the binary \textbf{classification} task. We randomly select around 82K and 18K images as the training and validation set. Then we adopt representation function $\lambda$ as pre-trained ResNet-18 \citep{he2016deep} and classifier $h$ as two-fully connected layers. We report the test-set sub-group average accuracy and sufficiency gap ($\Delta\text{Suf}_C$) in Tab.~\ref{tab:fair_class} and Fig.~\ref{fig:celeba}.



The results in the CelebA show similar behaviors with the Toxic comments. Specifically, the DP based fair approaches (\RN{2}, \RN{3}) did not effectively improve $\Delta\text{Suf}_C$, shown in Tab.~\ref{tab:fair_class}.  In contrast, the sufficiency can be significantly improved in baselines (\RN{4}, \RN{5}) and implicit approach without largely losing the accuracy. Specifically, Fig.~\ref{fig:celeba} visualizes the accuracy-fair trade-off curve, where the later three approaches show quite similar behaviors.

\begin{table*}[t]
\centering
\caption{Fair Regression. MSE and $\Delta\text{Suf}_R$ in Law dataset (left) and NLSY dataset (right)}
\label{tab:fair_regression}
\resizebox{0.45\textwidth}{!}{
\begin{tabular}{ccc}\toprule
     \textcolor{blue}{Law} & MSE~($\downarrow$) & $\Delta\text{Suf}_R$~($\downarrow$) \\ \midrule
     ERM~(\RN{1}) & 0.190 $\pm$ 0.005 & 0.160 $\pm$ 0.007  \\\midrule
     Adv\_debias~(\RN{2}) & 0.223 $\pm$ 0.008 & 0.188 $\pm$ 0.012 \\
     Mixup~(\RN{3})  & 0.216 $\pm$ 0.012 & 0.172 $\pm$ 0.007 \\ \midrule
     IRM\_v1~(\RN{4}) & 0.208 $\pm$ 0.006 & 0.096 $\pm$ 0.006  \\ 
     One\_step~(\RN{5}) & 0.204 $\pm$ 0.007 & 0.125 $\pm$ 0.010 \\
     Implicit & 0.198 $\pm$ 0.005  & 0.091 $\pm$ 0.011 \\ \bottomrule
     \end{tabular}
}
\resizebox{0.45\textwidth}{!}{
\begin{tabular}{ccc}\toprule
     \textcolor{blue}{NLSY} & MSE~($\downarrow$) & $\Delta\text{Suf}_R$~($\downarrow$) \\ \midrule
     ERM~(\RN{1}) &  1.939 $\pm$ 0.021 &  0.246 $\pm$ 0.019  \\\midrule
     Adv\_debias~(\RN{2}) & 1.982 $\pm$ 0.016 & 0.252 $\pm$ 0.020  \\
     Mixup~(\RN{3})  &  1.979 $\pm$ 0.025  &  0.246 $\pm$ 0.023 \\ \midrule
     IRM\_v1~(\RN{4}) &  1.927 $\pm$ 0.031 & 0.077 $\pm$ 0.009  \\  
     One\_step~(\RN{5}) & 1.904 $\pm$ 0.027 & 0.090 $\pm$ 0.019 \\
     Implicit &  1.906 $\pm$ 0.019  & 0.051 $\pm$ 0.005 \\ \bottomrule
     \end{tabular}
}
\end{table*}

\begin{figure*}[h]
  \centering
  \subfigure[ERM]{\label{fig:erm-law}\includegraphics[width=50mm]{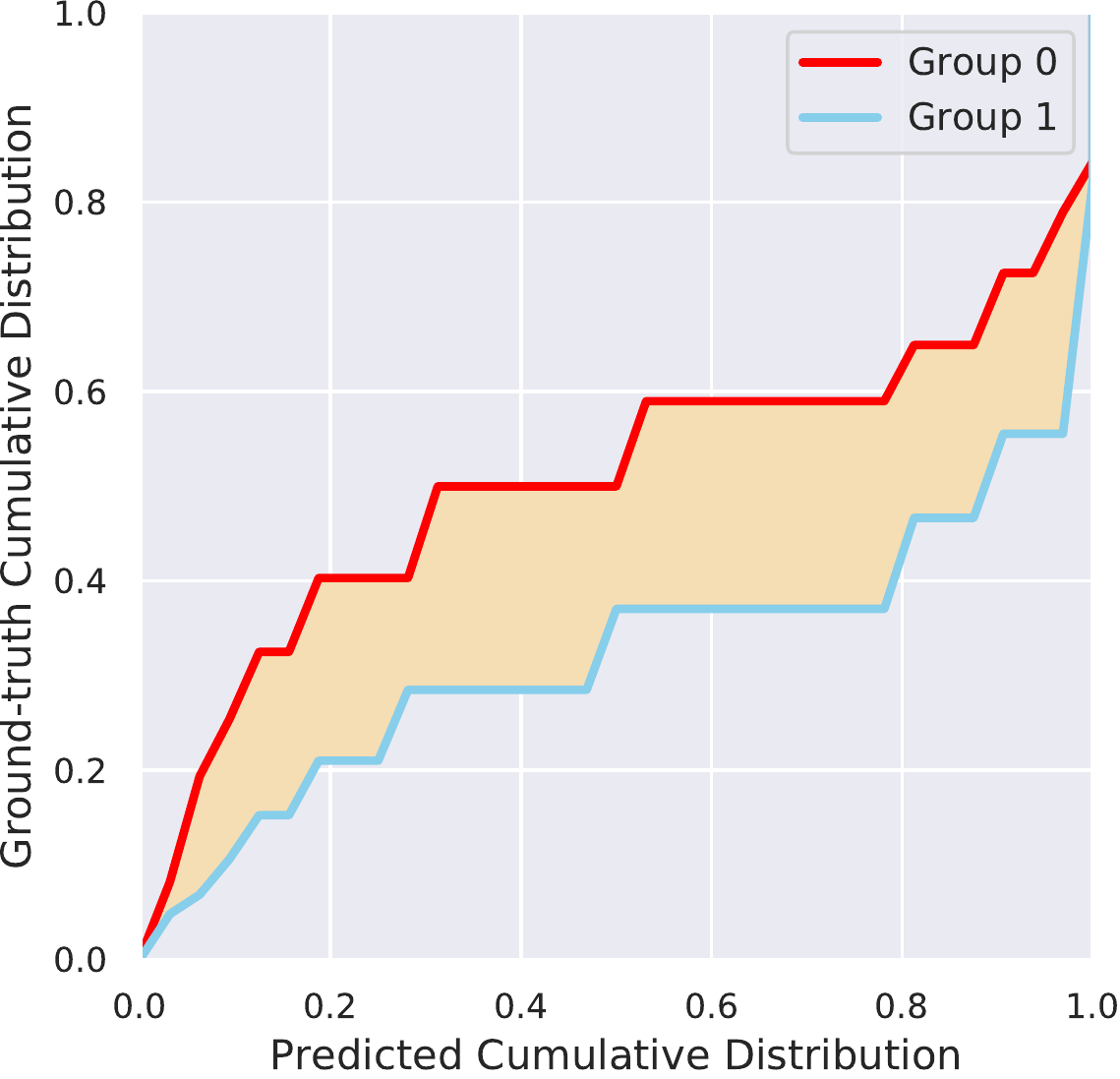}}
  \quad
  \subfigure[Fair Mix-up]{\label{fig:mix-up-law}\includegraphics[width=50mm]{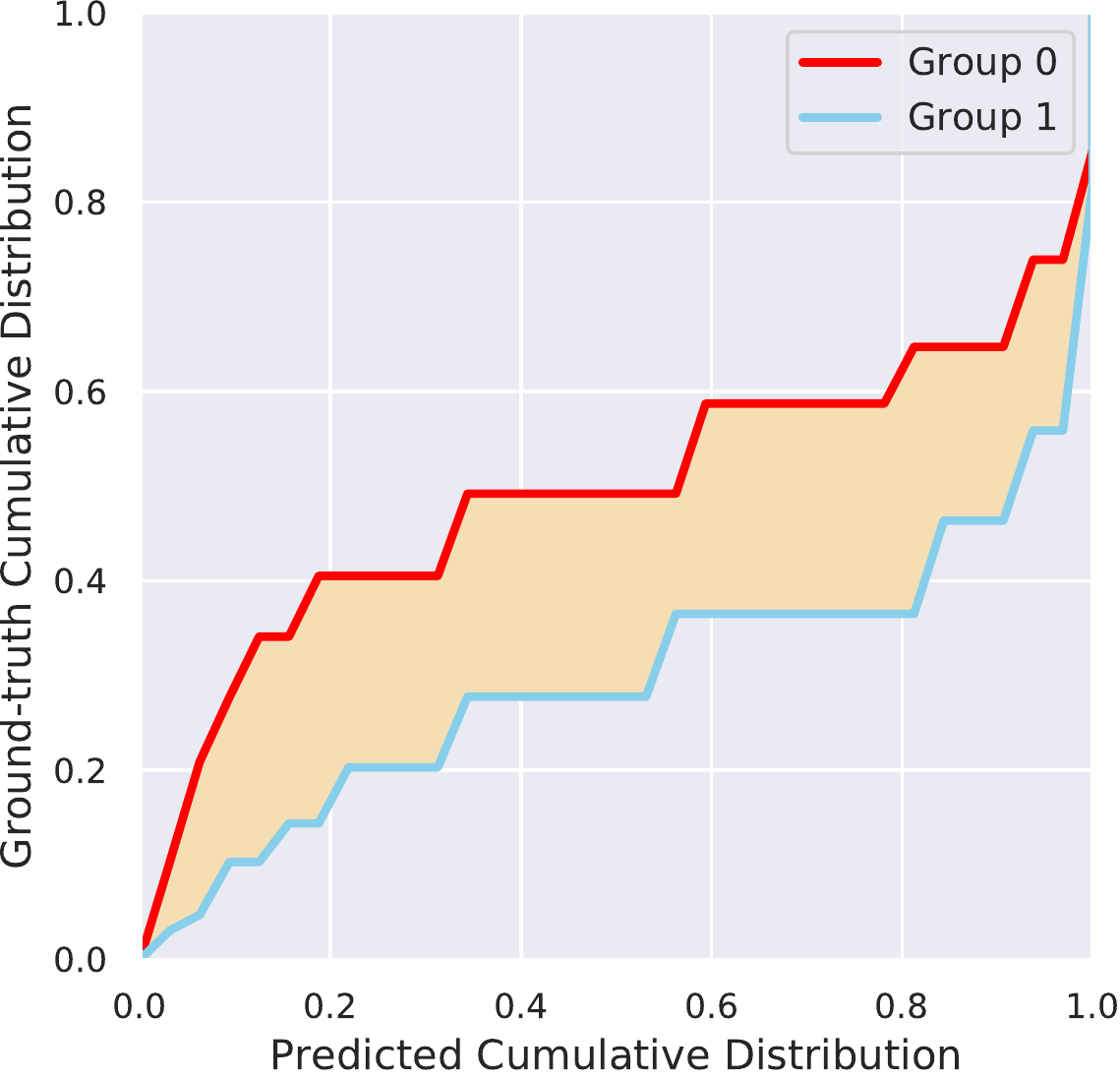}}
  \quad
  \subfigure[Implicit]{\label{fig:implict-law}\includegraphics[width=50mm]{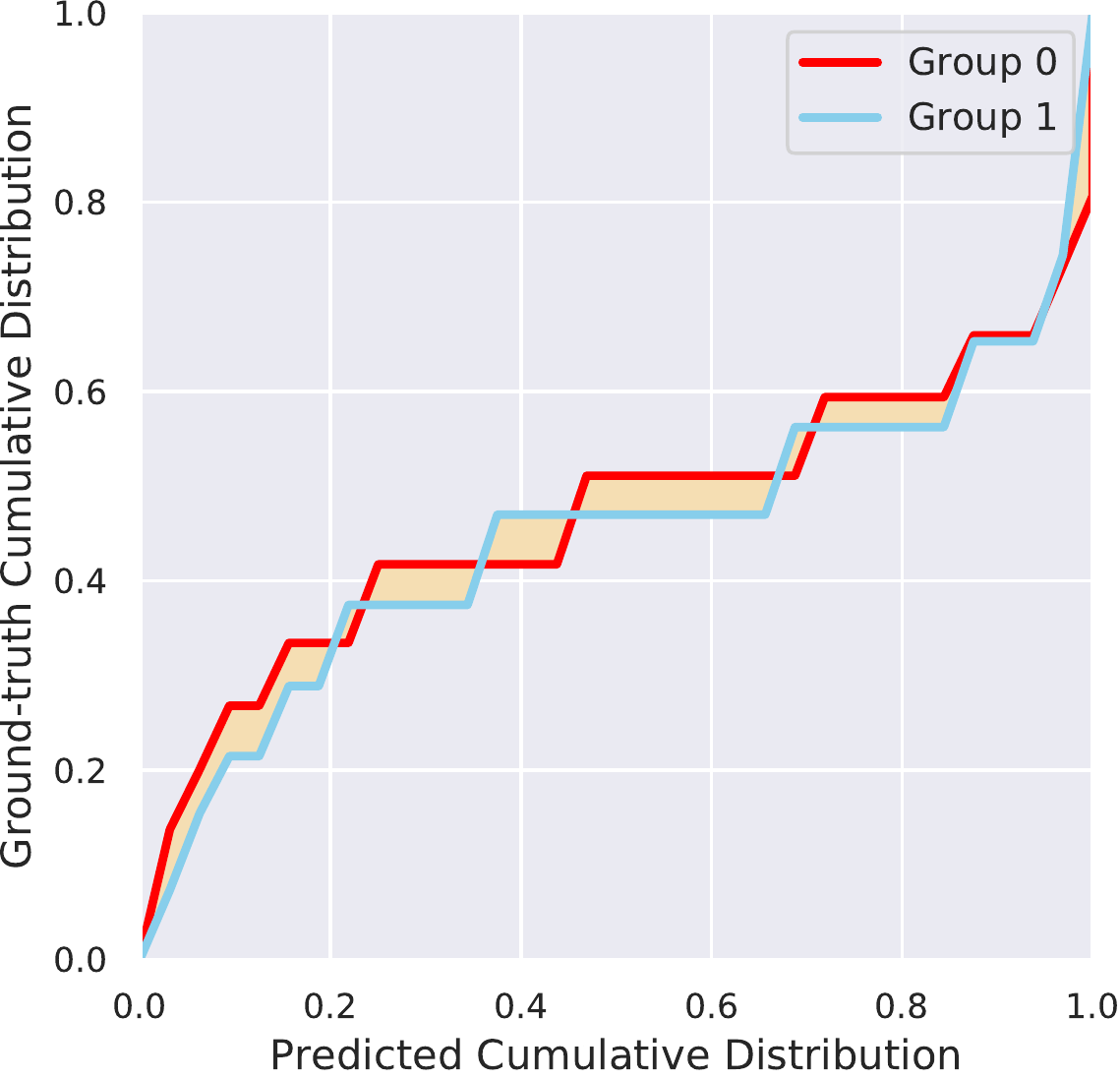}}
  \caption{Illustration of the sufficiency gap ($\Delta\text{Suf}_R$) in Law dataset (regression). The ERM and Fair mix-up suffer a high $\Delta\text{Suf}_R$, while the proposed implicit alignment can significantly mitigate the sufficiency gap. }
  \label{fig:sufficient_gap}
\end{figure*} 

\paragraph{Ablation: Computational benefits of Implicit Alignment} To show the efficiency of implicit approach in deep neural network, we empirically evaluated the computing time of $T$-inner step explicit alignment and implicit approach. The experimental results (shown in Fig.\ref{fig:ablation_toxic}) verified the computational efficiency of Implicit alignment. Notably, a large inner-optimization step does not considerably increase the whole computational time of implicit approach with different iterations of conjugated gradient solver. In contrast, the corresponding computational time complexity in explicit alignment linearly scales with the inner-optimization steps, which is consistent with our analysis.


\subsection{Law Dataset}
\begin{figure}[h]
    \centering
    \includegraphics[width=0.34\textwidth]{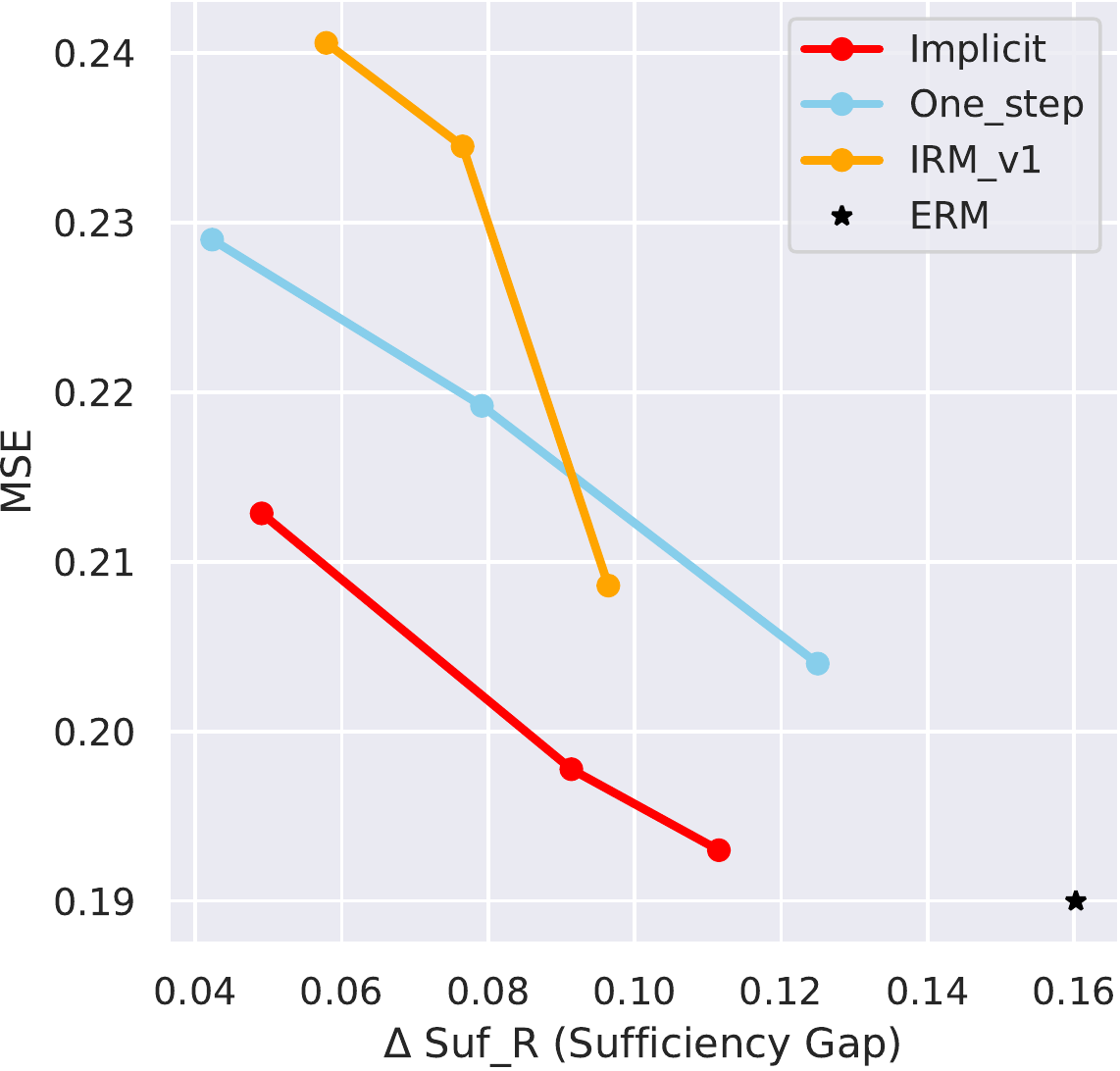}
    \caption{Law Dataset (regression). MSE-Fair Trade-off curve}\label{fig:law}
\end{figure}

The Law Dataset is a \textbf{regression} task to predict a students GPA (real value, ranging from $[0,4]$), where the data is utilized from the School Admissions Councils National Longitudinal Bar Passage Study \citep{Wightman1998} with 20K examples.  In the regression task, we adopt the square loss and \emph{race} as the protected feature (white versus non-white). We adopt  $\lambda$ as the one fully connected layer with hidden dimension 100 and Relu activation and predictor $h$ as a linear predictor. We report the test-set sub-group average MSE (Mean Square Error) and sufficiency gap ($\Delta\text{Suf}_R$) in Tab.~\ref{tab:fair_regression} and Fig.~\ref{fig:law}.

    
Compared to the classification task, the results show similar behaviors in the regression. Specifically, the DP based fair approaches (\RN{2}, \RN{3}) still increase $\Delta\text{Suf}_R$ in the regression. In contrast, the gap is significantly improved in our proposed approach and baseline (\RN{4},\RN{5}). Specifically, Fig.~\ref{fig:sufficient_gap} visualizes the sufficiency-gap of different approaches, where the implicit approach significantly mitigates the sufficiency gap. 
In addition, Fig.~\ref{fig:law} describes the MSE-sufficiency gap curve, which further justifies the benefits of implicit approach with a better trade-off between the prediction performance and fairness. 

\subsection{NLSY Dataset}
\begin{figure}[h]
    \centering
    \includegraphics[width=0.34\textwidth]{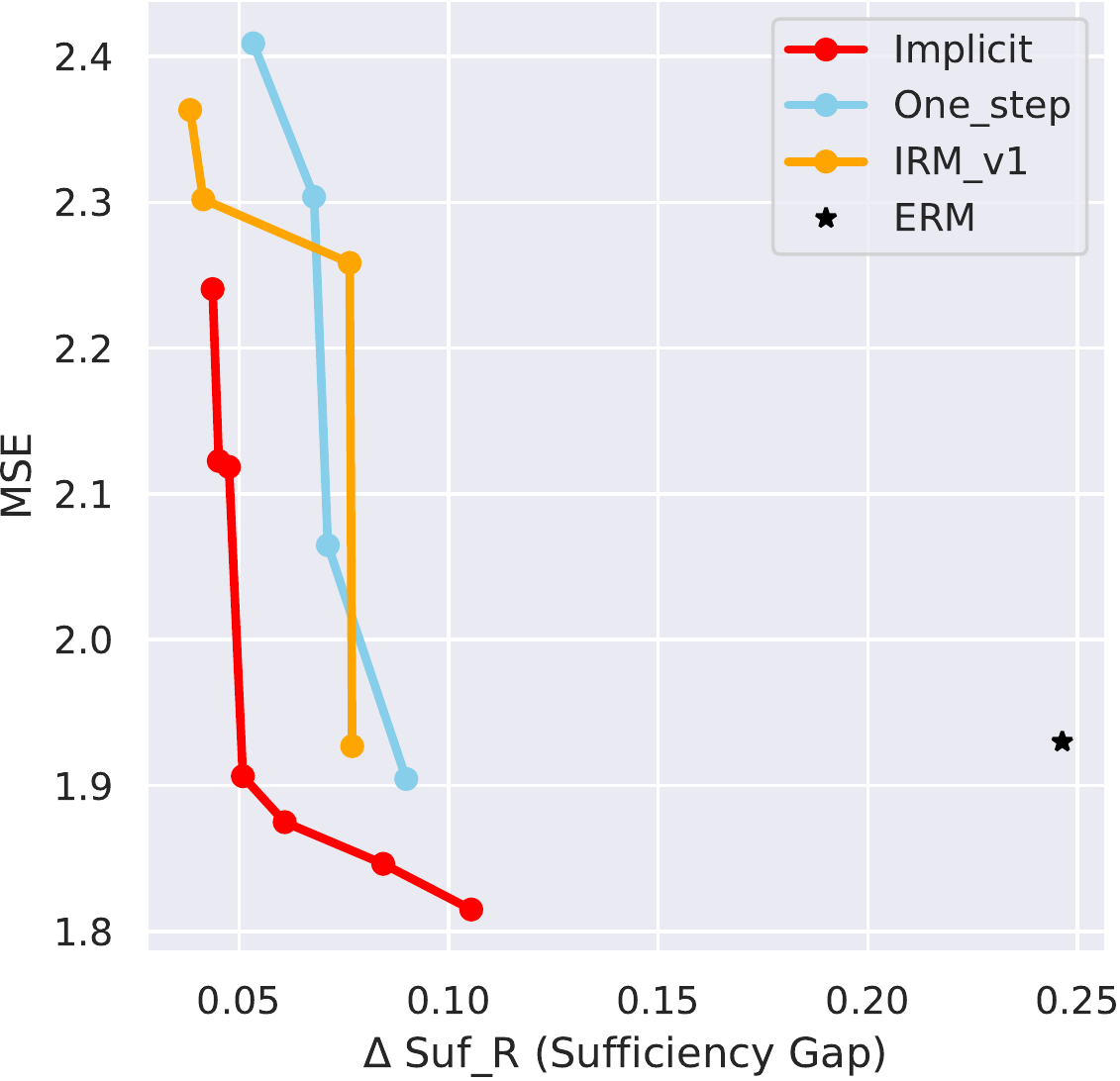}
    \caption{NLSY (regression). MSE-Fair Trade-off curve}\label{fig:8}
 \end{figure}

The National Longitudinal Survey of Youth \citep{NLSY} dataset is a \textbf{regression} task with around 7K dataset, which involves the survey results of the U.S. Bureau of Labor Statistics. It is intended to gather information on the labor market activities and other life events of several groups for predicting the income $y$ of each person. We treat the \emph{gender} as the protected feature. We also normalize the output $y$ by diving the $10,000$, then the final output $y$ ranges around $[0,8]$. The prediction loss is also the square loss. We adopt representation $\lambda$ as the two fully connected layers with hidden dimension 200 and Relu activation and predictor $h$ as a linear predictor. We report the testset sub-group average MSE (Mean Square Error) and Sufficiency Gap ($\Delta\text{Suf}_R$) in Tab.~\ref{tab:fair_regression} and Fig.~\ref{fig:8}.

Tab.~\ref{tab:fair_regression} provides similar trends with other datasets. Baselines (\RN{4},\RN{5}) and implicit approach effective control the sufficiency gap, while the DP based approach generally fails to improve the gap. Fig.~\ref{fig:8} reveals a slightly better approximation-fair trade off for the implicit approach. Finally, Fig.~\ref{fig:sufficient_gap_nlsy} (in Appendix) visualizes the sufficiency gap of different algorithms. The gap is actually significantly improved while the calibration gap still exists, which is consistent with \citep{pmlr-v97-liu19f}. Therefore it can be quite interesting and promising to analyze the triple trade-off between the sufficiency gap, probabilistic calibration and prediction performance in the regression in the future.


\section{Conclusion}
We considered the fair representation learning from a novel perspective through encouraging the invariant optimal predictors on the top of data representation. We formulated this problem as a bi-level optimization and proposed an implicit alignment algorithm. We further demonstrated the bi-level objective is to fulfil the Sufficiency rule. Then we analyzed the error gap of the implicit algorithm, which reveal the trade-off of biased gradient approximation and fairness constraints. The empirical results in both classification and regression settings suggest the consistently improved fairness measurement. Finally, we think the future work can include developing computationally efficient explicit algorithms for avoiding the biased gradient computation.

\section*{Limitations}
We considered a novel fair representation learning perspective to encourage the sufficiency rule. Simultaneously this work remains several limitations.

In the proposed algorithm, we need a two-step optimization with tolerance $\epsilon$ and $\delta$. As for controlling $\epsilon$ (the tolerance w.r.t. the predictor $h$), since $h$ is a shallow network with one or two layers, then optimizing over $h$ will be relatively easy. As for $\delta$, since the representation $\lambda$ could be highly non-convex and high-dimensional, controlling $\delta$ would be quite difficult in theory. In practice, we generally control the steps in the conjugate gradient while it is unclear the convergence behavior in the highly non-convex settings.

The current paper mainly focus the binary sensitive attribute with two subgroups. Although it is feasible to extend the multi-attribute settings by consider the pair-wise path alignment, but it would be promising to consider an effect algorithm for the multi-attribute.

The performance and fair trade-off is induced by the efficient gradient estimation in the bi-level objective. Thus it would be promising to develop an efficient explicit path approach for avoiding such a trade-off.

\section*{Acknowledgments}
The authors appreciate the constructive feedback and suggestions from anonymous Reviewers and Meta-Reviewers. The authors also would like to thank Gezheng Xu and Jun Xiao for the discussion and proof-reading the manuscript.  

C. Shui and C. Gagné acknowledge support from NSERC-Canada and CIFAR.  B. Wang, J. Li are supported by NSERC Discovery Grants Program. 

\bibliography{iclr2022_conference}
\bibliographystyle{icml2022}

\newpage
\appendix
\onecolumn
\section{Sufficiency rule: comparison with other fair criteria}
\paragraph{Sufficiency v.s. Independence}
We will demonstrate:

If $\D_0(Y=y)\neq \D_1(Y=y)$ (i.e, different label distribution in the sub-groups), the Sufficiency and Independence rule cannot both hold. 

\begin{proof}
Since we consider the binary-classification with $\calY=\{-1,1\}$, the expectation and conditional can be expressed as the probability of predicted output.
\[
\D_0(\hat{Y}=1) = \frac{1}{2}(1+\E_{\D_0}[\hat{Y}]) ,~~\D_0(Y=1|\hat{Y}=t) = \frac{1}{2}(1+\E_{\D_0}[Y|\hat{Y}=t])
\]
Then sufficiency and independence are equivalent to: $\D_0(\hat{Y}=t)=\D_1(\hat{Y}=t)$ and $\D_0(Y=y|\hat{Y}=t) =\D_1(Y=y|\hat{Y}=t)$ both hold for any $t,y$. Then the joint distribution of $\hat{Y},Y$ should be identical:
\[
\D_0(\hat{Y}=t,Y=y)=\D_1(\hat{Y}=t,Y=y), ~~\forall t,y
\]
Then the marginal distribution $\D_0(Y=y)=\D_1(Y=y)$ must holds. 

If $\D_0(Y=y)\neq\D_1(Y=y)$, the joint distribution is not equal:
\[
\D_0(\hat{Y}=t,Y=y)\neq\D_1(\hat{Y}=t,Y=y), ~~\forall t,y
\]
Since $\D_0(\hat{Y}=t,Y=y) = \D_0(Y=y|\hat{Y}=t) \D_0(\hat{Y}=t)$, thus either $\D_0(\hat{Y}=t)\neq \D_1(\hat{Y}=t)$ or $\D_0(Y=y|\hat{Y}=t)\neq\D_1(Y=y|\hat{Y}=t)$ must hold for at least one term. I.e, the sufficiency and Independence could not both hold.
\end{proof}

\paragraph{Sufficiency v.s. Separation}
We will demonstrate:

If $\D_0(Y=y)\neq \D_1(Y=y)$ and joint distribution of $(Y,\hat{Y})$ has positive probability in $\D_0,\D_1$, the Sufficiency and Separation rule cannot both hold. 

\begin{proof}
Based on the previous results, if $\D_0(Y=y)\neq \D_1(Y=y)$, then the joint distribution of $(Y,\hat{Y})$ are not identical:
\[
\D_0(\hat{Y}=t,Y=y)\neq\D_1(\hat{Y}=t,Y=y), ~~\forall t,y
\]
Then either $\D_0(\hat{Y}=t)\neq \D_1(\hat{Y}=t)$ or $\D_0(Y=y|\hat{Y}=t)\neq\D_1(Y=y|\hat{Y}=t)$ must hold for at least one term (conclusion of previous result).

If the sufficiency and separation both hold when $\D_0(Y=y)\neq \D_1(Y=y)$, it must be the following case:
\begin{align*}
    & \D_0(\hat{Y}=t)\neq \D_1(\hat{Y}=t), ~~~ \D_0(Y=y|\hat{Y}=t)=\D_1(Y=y|\hat{Y}=t)\\
    & \D_0(Y=y)\neq \D_1(Y=y), ~~~
    \D_0(\hat{Y}=t|Y=y)=\D_1(\hat{Y}=t|Y=y)
\end{align*}
However, we will prove it is impossible in the classification. Based on Bayes rule:
\[
\D_0(Y=y|\hat{Y}=t) = \frac{\D_0(\hat{Y}=t|Y=y)\D_0(Y=y)}{\D_0(\hat{Y}=t)} = \frac{\D_1(\hat{Y}=t|Y=y)\D_1(Y=y)}{\D_1(\hat{Y}=t)} = \D_1(Y=y|\hat{Y}=t)
\]
Thus we should have 
\[
\frac{\D_0(Y=y)}{\D_0(\hat{Y}=t)} = \frac{\D_1(Y=y)}{\D_1(\hat{Y}=t)}, ~~\forall t,y
\]
We consider the binary classification by denoting $\D_0(Y=1)=p,\D_1(y=1)=q, \D_0(\hat{Y}=1)=\hat{p}, \D_1(\hat{Y}=1)=\hat{q}$, then we have:
\[
\frac{p}{\hat{p}} = \frac{q}{\hat{q}}, \frac{1-p}{\hat{p}} = \frac{1-q}{\hat{q}}, \frac{p}{1-\hat{p}} = \frac{q}{1-\hat{q}}, \frac{1-p}{1-\hat{p}} = \frac{1-q}{1-\hat{q}}
\]
There exists the unique non-zero solution of $p=q=\hat{p}=\hat{q}=0.5$, which clearly contradicts our assumptions. 
\end{proof}

\subsection{Comparison Tables}
For the sake of completeness, we list common fair criteria for the comparison.
\begin{table}[!ht]
    \centering
    \caption{Common Fair criteria in classification}
    \resizebox{\columnwidth}{!}{
    \begin{tabular}{l|l|l|l|l}
    \toprule
        ~ & General Definition & Binary classification & Definition  & Relation  \\ \midrule
        Independence & $\D_0(\hat{Y})=\D_1(\hat{Y})$ & Demographic parity & $\D_0(\hat{Y}=1)=\D_1(\hat{Y}=1)$ & Equivalent  \\ \midrule
        Separation   & $\D_0(\hat{Y}|Y=y) = \D_1(\hat{Y}|Y=y),\forall y$ & Equalized odds & $\D_0(\hat{Y}=1|Y=y) = \D_1(\hat{Y}=1|Y=y),\forall y$ & Equivalent \\ \midrule
        Separation   & ~ & Equal opportunity & $\D_0(\hat{Y}= 1 |Y= 1) = \D_1(\hat{Y}=1|Y=1)$ &  Relaxation \\ \midrule
        Sufficiency  & $\D_0(Y|\hat{Y}=y) = \D_1(Y|\hat{Y}=y),\forall y$ & Conditional use accuracy equality & $\D_0(Y = y|\hat{Y}=y) = \D_1(Y = y|\hat{Y}=y),\forall y$ & Equivalent  \\ \midrule
        Sufficiency  & ~ & Predictive parity  & $\D_0(Y = 1 |\hat{Y}=1) = \D_1(Y = 1 |\hat{Y}=1)$ & Relaxation  \\ 
    \bottomrule
    \end{tabular}}
\end{table}

\section{Proof Proposition 3.1}
We consider the regression and classification separately.
\paragraph{Regression} According to the definition, given a fixed and deterministic representation $\lambda$, we have 
\[
\calL_0(h,\lambda) = \E_{\D_0}(h(z)-y)^2
\]
Since it is a functional optimization w.r.t. the function $h$, through using the calculus of variations \cite{mse_opt}, 
\[
\frac{\delta \calL_0(h,\lambda)}{\delta h(z)} = 2 \int [h(z)-y] \D_0(z,y) dy = 0
\]
Solving for $h(z)$, and using the sum and product rules of probability, we obtain
\[
h^{\star}_{0}(z) = \frac{\int y \D_0(z,y) dy}{\D_0(z)} = \int y \D_0(y|Z=z) dy = \E_{\D_0}[Y|Z=z]
\]
Then we have $h_{0}^{\star}(z) = \E_{\D_0}[Y|Z=z]$. As for $\D_1$, we apply the same strategy with $h_{1}^{\star}(z) = \E_{\D_1}[Y|Z=z]$. Based on the invariant optimal predictor, we have 
$\E_{\D_0}[Y|Z=z]=\E_{\D_1}[Y|Z=z]$ with $z=\lambda(x)$.




\paragraph{Classification} According to the definition, we have:
\[
\calL_0(h,\lambda) = \E_{\D_0} \log(1+\exp(-yh(z)))
\]
Since the optimal predictor on the logistic loss is the log-conditional density ratio: $h_0^{\star}(z) = \log\left(\frac{\D_0(Y=1|Z=z)}{\D_0(Y=-1|Z=z)}\right)$. Observe that in the binary classification with $Y=\{-1,1\}$, we have $\D_0(Y=1|Z=z)=\frac{1}{2}(1+\E_{\D_0}[Y|Z=z])$ and  $\D_0(Y=-1|Z=z)=\frac{1}{2}(1-\E_{\D_0}[Y|Z=z])$, then we have:
\[h_0^{\star}(z) = \log\left(\frac{1+\E_{\D_0}[Y|Z=z]}{1-\E_{\D_0}[Y|Z=z]}\right)\]
As for $\D_1$, we adopt the same strategy and we have $\log\left(\frac{1+\E_{\D_0}[Y|Z=z]}{1-\E_{\D_0}[Y|Z=z]}\right) = \log\left(\frac{1+\E_{\D_1}[Y|Z=z]}{1-\E_{\D_1}[Y|Z=z]}\right)$, then we have $\E_{\D_0}[Y|Z=z] = \E_{\D_1}[Y|Z=z]$.

As for the predictive parity, since we have $\E_{\D_0}[Y|Z=z]=\E_{\D_1}[Y|Z=z]$ and $h^{\star} = h_1^{\star} = h_2^{\star}$, then we have $\E_{\D_0}[Y|h^{\star}(z)]=\E_{\D_1}[Y|h^{\star}(z)]$.

\section{Approximation Error}\label{appendix:approx_error}
\begin{theorem}[Approximation Error Gap] Suppose that (1) \textbf{Smooth Predictive Loss}. The first-order derivatives and second-order derivatives of $\calL$ are Lipschitz continuous; (2) \textbf{Non-singular Hessian matrix}. We assume $\nabla_{h_0,h_0}\calL_0(h_0,\lambda), \nabla_{h_1,h_1}\calL_1(h_1,\lambda)$, the Hessian matrix of the inner optimization problem, are invertible. (3) \textbf{Bounded representation and predictor function}. We assume the $\lambda$ and $h$ are bounded, i.e., $\|\lambda\|, \|h\|$ are upper bounded by the predefined positive constants. Then the approximation error between the ground truth and algorithmic estimated gradient w.r.t. the representation is be upper bounded by:
\[\|\text{grad}(\lambda)- \tilde{\text{grad}}^{\delta}(\lambda)\| = \mathcal{O}(\kappa\epsilon + \epsilon + \delta).\]
\end{theorem}
\begin{proof}
We denote $\text{grad}(\lambda)$ as the ground truth gradient w.r.t. $\lambda$ in outer-loop loss (given the optimal predictor $h_0^{\star}$, $h_1^{\star}$). Then we aim to bound 
\[\|\text{grad}(\lambda)- \tilde{\text{grad}}^{\delta}(\lambda) \|\]
We first introduce the following terms for facilitating the proof:
\begin{align*}
   & A_0^{\epsilon} = \nabla_{h_0}\nabla_{\lambda} \calL_0(h_0^{\epsilon},\lambda), A_1^{\epsilon} = \nabla_{\lambda}\nabla_{h_1} \calL_1(h_1^{\epsilon},\lambda),
    A_0^{\star} = \nabla_{\lambda}\nabla_{h_0} \calL_0(h_0^{\star},\lambda), A_1^{\star} = \nabla_{\lambda}\nabla_{h_1} \calL_1(h_1^{\star},\lambda),\\
   & B_0^{\epsilon} = \nabla_{\lambda} \calL_0(h_0^{\epsilon},\lambda),  B_1^{\epsilon} = \nabla_{\lambda} \calL_1(h_1^{\epsilon},\lambda),
   B_0^{\star} = \nabla_{\lambda} \calL_0(h_0^{\star},\lambda),  B_1^{\star} = \nabla_{\lambda} \calL_1(h_1^{\star},\lambda),\\
   & \mathbf{p}_0^{\star} = \left(\nabla^2_{h_0} \calL_0(h_0^{\star},\lambda) \right)^{-1} \left(\nabla_{h_0}\calL_0(h_0^{\star},\lambda) + \kappa(h^{\star}_0-h^{\star}_1)\right),\\
   & \mathbf{p}_1^{\star} = \left(\nabla^2_{h_1} \calL_1(h_1^{\star},\lambda) \right)^{-1} \left(\nabla_{h_1}\calL_1(h_1^{\star},\lambda) - \kappa(h^{\star}_0-h^{\star}_1)\right).
\end{align*}
Then the approximation error gap can be expressed as:
\begin{align*}
    \|\text{grad}(\lambda)- \tilde{\text{grad}}^{\delta}(\lambda) \| & = \|\left(B_0^{\star} - A_0^{\star}\mathbf{p}_0^{\star} + B_1^{\star} - A_1^{\star}\mathbf{p}_1^{\star}\right) - \left(B_0^{\epsilon} - A_0^{\epsilon}\mathbf{p}_0^{\delta} + B_1^{\epsilon} - A_1^{\epsilon}\mathbf{p}_1^{\delta} \right)\|\\
    & \leq \sum_{i=0}^{1} \|B_i^{\star} -B_i^{\epsilon}\| + \sum_{i=0}^1 \|A_i^{\star}\mathbf{p}_i^{\star} - A_i^{\delta}\mathbf{p}_i^{\delta}\|
\end{align*}
Due to the symmetric of $\D_0$ and $\D_1$, we only focus on the term on $i=0$, the the upper bound in $i=1$ can be derived analogously.

As for bounding $\|B_0^{\star} -B_0^{\epsilon}\|$,  since we assume first order derivative of the loss is Lipschitz functions (with constant $L_1$), then we have :
\[\|B_0^{\star} -B_0^{\epsilon}\| \leq L_1\|h_0^{\star}-h_0^{\epsilon}\|\leq \epsilon L_1\]
Then the second term can be upper bounded by three terms:
\[ \|A_0^{\star}\mathbf{p}_0^{\star} - A_0^{\delta}\mathbf{p}_0^{\delta}\|\leq \underbrace{\|A_0^{\star}\mathbf{p}_0^{\star} - A_0^{\star}\mathbf{p}_0\|}_{(1)} + \underbrace{\|A_0^{\star}\mathbf{p}_0 - A_0^{\epsilon}\mathbf{p}_0\|}_{(2)} + \underbrace{\|A_0^{\epsilon}\mathbf{p}_0 - A_0^{\epsilon}\mathbf{p}^{\delta}_0\|}_{(3)} \]
Before estimating the upper bound, we first demonstrate $\|A_0^{\epsilon}\|$ and $\|A_0^{\star}\|$ are also bounded.

Since we assume $\lambda$ and $h$ are bounded (assuming the bounded constant as $\eta$ and $\phi$), the second order derivative are Lipschitz (with constant $L_2$). Then we consider another fixed point $(\lambda^{\prime}, h_0^{\star}(\lambda^{\prime}))$ with bounded second order derivative: $A_0 = \nabla^2_{h_0,\lambda} \calL_0(h_0^{\star}(\lambda^{\prime}),\lambda^{\prime})$ and $\|A_0\|\leq A$. We have:
\[
\|A_{0}^{\star} - A_0\|_2 \leq L_2 \|[h_0^{\star}(\lambda),\lambda]- [h_0^{\star}(\lambda^{\prime}),\lambda^{\prime}]\|_2 \leq L_2 \sqrt{\eta^2+\phi^2}
\]
Thus we have $\|A_{0}^{\star}\|\leq A + L_2 \sqrt{\eta^2+\phi^2} = A_{\sup}^{\star}$. As for the second derivative at point $h_0^{\epsilon}$, it can be upper bounded analogously with a similar constant $A_{\sup}^{\epsilon}$. 

\paragraph{The upper bound of term (1)} We have:
\[
\|A_0^{\star}\mathbf{p}_0^{\star} - A_0^{\star}\mathbf{p}_0\| \leq \|A_0^{\star}\| \|\mathbf{p}_0^{\star}-\mathbf{p}_0\|
\]
We have proved $\|A_0^{\star}\|$ is upper bounded by $A_{\sup}^{\star}$. We additionally introduce the following auxiliary terms:
\begin{align*}
  & P_0^{\star} = \left(\nabla^2_{h_0} \calL_0(h_0^{\star},\lambda) \right)^{-1}, P_0^{\epsilon} = \left(\nabla^2_{h_1} \calL_1(h_1^{\star},\lambda) \right)^{-1}.\\
  & b_0^{\star} = \nabla_{h_0}\calL_0(h_0^{\star},\lambda) + \kappa(h^{\star}_0-h^{\star}_1), b_0^{\epsilon} = \nabla_{h_0}\calL_0(h_0^{\epsilon},\lambda) + \kappa(h^{\epsilon}_0-h^{\epsilon}_1)
\end{align*}
Then we have:
\begin{align*}
    \|\mathbf{p}_0^{\star}-\mathbf{p}_0 \| & =  \| P_0^{\star}b_0^{\star}- P_0^{\epsilon}b_0^{\epsilon}\|\\
    & \leq \|P_0^{\star}b_0^{\star}-P_0^{\star}b_0^{\epsilon}\| + \|P_0^{\star}b_0^{\epsilon}- P_0^{\epsilon}b_0^{\epsilon}\|\\
    & \leq \|P_0^{\star}\| \|b_0^{\star}-b_0^{\epsilon}\| + \|b_0^{\epsilon}\|\|P_0^{\star}-P_0^{\epsilon}\|
\end{align*}
As for the $\|P_{0}^{\star}\|$, since we assume the Hessian matrix is invertible thus its norm is upper bounded by some constant (denoted as $A_{-1}$).  As for $\|b_0^{\star}-b_0^{\epsilon}\|$, we have:
\begin{align*}
    \|b_0^{\star}-b_0^{\epsilon}\| & \leq \|\nabla_{h_0}\calL_0(h_0^{\star},\lambda)-\nabla_{h_0}\calL_0(h_0^{\epsilon},\lambda)\| + 2\kappa\epsilon \\
    & \leq \epsilon L_1 + 2\kappa\epsilon 
\end{align*}
Thus we have $\|P_0^{\star}\| \|b_0^{\star}-b_0^{\epsilon}\|\leq A_{-1} (\epsilon L_1 + 2\kappa\epsilon)$. 

As for $\|b_0^{\epsilon}\|$, we can easily verify that it is indeed bounded by some constant $b$. For the first term, we can adopt the same strategy in proving bounded $\|A_0^{\star}\|$. As for the second term in $b_0^{\epsilon}$, it is upper bounded by $2\kappa\phi$, due to the bounded predictor.

We now demonstrate $\|P_0^{\star}-P_0^{\epsilon}\|$. Denoting $\Delta= (P_0^{\star})^{-1} - (P_0^{\epsilon})^{-1}$, then according to the second order Lipschitz assumption, we have: $\|\Delta\|\leq \epsilon L_2$. Plugging in the result, we have:
\begin{align*}
    \|P_0^{\star}-P_0^{\epsilon}\| = \| (P_0^{\star})\Delta(P_0^{\epsilon})\| \leq \| P_0^{\star}\|\|\Delta\|\|P_0^{\epsilon}\| \leq (A_{-1})^2L_2 \epsilon
\end{align*}
We still adopt the assumption that the bounded Hessian-inverse matrix by $A_{-1}$. 

Plugging in all the results, we have:
\[
(1)\leq A_1 (\epsilon L_1 + 2\kappa\epsilon) + b (A_1)^2L_2 \epsilon :=\mathcal{O}(\kappa\epsilon+\epsilon)
\]

\paragraph{The upper bound of term (2)} We have:
\[
\|A_0^{\star}\mathbf{p}_0 - A_0^{\epsilon}\mathbf{p}_0\|\leq \|\mathbf{p}_0\|_2 \| A_0^{\star}-A_0^{\epsilon}\|
\]
Since we assume the loss is second-order Lipschitz, thus we have
\begin{align*}
    \| A_0^{\star}-A_0^{\epsilon}\| = \|\nabla_{\lambda}\nabla_{h_0} \calL_0(h_0^{\star},\lambda)-\nabla_{\lambda}\nabla_{h_0} \calL_0(h_0^{\epsilon},\lambda)\|\leq L_2 \|h_0^{\star}-h_0^{\epsilon}\| \leq \epsilon L_2
\end{align*}
We can also demonstrate $\|\mathbf{p}_0\|$ is bounded. According to the definition we have:
\begin{align*}
    \|\mathbf{p}_0\|& \leq \|\left(\nabla^2_{h_0} \calL_0(h_0^{\epsilon},\lambda) \right)^{-1}\| \|\left(\nabla_{h_0}\calL_0(h_0^{\epsilon},\lambda) + \kappa(h^{\epsilon}_0-h^{\epsilon}_1)\right)\|\\
    & \overset{(i)}{\leq} A_{-1}(L_1 \|h_0^{\star}-h_0^{\epsilon}\|_2 + 2\kappa\phi)\\
    & \overset{(ii)}{\leq} A_{-1}(\epsilon L_1 + 2\kappa\phi)
\end{align*}
For (i), we assume: 1) the Hessian matrix is invertible thus its norm is surely upper bounded by some constant (denoted as $A_{-1}$), 2) the first-order derivative is Lipschitz (bounded by $L_1$), 3) the predictor $h$ is bounded. For (ii), we adopt the definition of $h_0^{\epsilon}$.

Therefore, the upper bound for Term (2) is formulated as:
\[
(2) \leq \epsilon L_2 A_{-1}(\epsilon L_1 + 2\kappa\phi) := 
\mathcal{O}(\kappa \epsilon)
\]

\paragraph{The upper bound of term (3)} We have:
\[
\|A_0^{\epsilon}\mathbf{p}_0 - A_0^{\epsilon}\mathbf{p}^{\delta}_0\|\leq \|A_0^{\epsilon}\| \|\mathbf{p}_0-\mathbf{p}^{\delta}_0\| \leq \delta A_{\sup}^{\epsilon} =\mathcal{O}(\delta)
\]

Through the upper bound in (1)-(3), we finally have the error between the estimated and ground-truth gradient:
\[
\|\text{grad}(\lambda)- \tilde{\text{grad}}^{\delta}(\lambda)\| = \mathcal{O}(\kappa\epsilon + \epsilon + \delta)
\]
\end{proof}

\section{The Convergence Behavior}\label{appendix:glo_convergence}
For the sake of completeness, we provide the convergence analysis of the proposed algorithm. 
\begin{proposition}
We execute the implicit alignment algorithm (Algo. 1), obtaining a sequence of  $\lambda_{1},\dots,\lambda_{k},\dots$. Supposing the fair constraint $\kappa$ is fixed. The optimization tolerances are summable: $\sum_{k}\epsilon^2_k \leq +\infty$ and $\sum_{k}\delta^2_k \leq +\infty$, then $\lambda_{k}$ is proved to be converged with \[\lim_{k\to\infty}\lambda_{k} = \lambda^{\star}.\] 
If the stationary point $\lambda^{\star}$ is also within the bounded norm, then we have:
\[
\text{grad}(\lambda^{\star}) = 0.
\]
\end{proposition}
\begin{proof}
We denote the entire outer-loop loss w.r.t. $\lambda$ as $\calL(\lambda)$, by the assumption the $\beta$-smooth loss $\calL$. Then at iteration $k+1$ and $k$, we have:
\begin{align*}
\calL(\lambda_{k+1}) & \leq \calL(\lambda_{k}) - \text{grad}(\lambda_k)^{T}(\lambda_k-\lambda_{k+1}) + \frac{\beta}{2}\|\lambda_{k+1}-\lambda_{k}\|^2 \\
& = \calL(\lambda_{k}) - \left(\text{grad}(\lambda_k)- \tilde{\text{grad}}^{\delta}(\lambda_k) + \tilde{\text{grad}}^{\delta}(\lambda_k) \right)^{T}(\lambda_k-\lambda_{k+1}) + \frac{\beta}{2}\|\lambda_{k+1}-\lambda_{k}\|^2 \\
& = \calL(\lambda_{k}) - \left(\text{grad}(\lambda_k)- \tilde{\text{grad}}^{\delta}(\lambda_k)\right)^{T}(\lambda_k-\lambda_{k+1}) - \tilde{\text{grad}}^{\delta}(\lambda_k)(\lambda_k-\lambda_{k+1}) +\frac{\beta}{2}\|\lambda_{k+1}-\lambda_{k}\|^2
\end{align*}
Since we assume the representation is within the bounded norm, the projection onto the convex set are non-expansive operators~\citep{boyd2004convex}. Then for any point $p,q$, we have $\|\text{proj}(p)-\text{proj}(q)\|^2\leq (p-q)^T\left(\text{proj}(p)-\text{proj}(q)\right)$. Then we set $\lambda_k$ and $\lambda_{k+1} = \lambda_{k} - \frac{1}{\beta}\tilde{\text{grad}}^{\delta}(\lambda_k)$, we have:
\[ \|\lambda_{k} - \lambda_{k+1}\|^2 \leq \frac{1}{\beta}(\tilde{\text{grad}}^{\delta}(\lambda_k))^T(\lambda_{k} - \lambda_{k+1})
\]
Plugging into the results, we have:
\begin{align*}
    \calL(\lambda_{k+1}) & \leq \calL(\lambda_{k}) - \left(\text{grad}(\lambda_k)- \tilde{\text{grad}}^{\delta}(\lambda_k)\right)^{T}(\lambda_k-\lambda_{k+1})-\frac{\beta}{2}\|\lambda_{k+1}-\lambda_{k}\|^2 \\
    & \leq \calL(\lambda_{k}) + \|\text{grad}(\lambda_k) - \tilde{\text{grad}}^{\delta}(\lambda_k)\|\|\lambda_k-\lambda_{k+1}\|-\frac{\beta}{2}\|\lambda_{k+1}-\lambda_{k}\|^2
\end{align*}
Rearranging the inequality, we have:
\[
\frac{\beta}{2}\|\lambda_{k+1}-\lambda_{k}\|^2 - \|\text{grad}(\lambda_k) - \tilde{\text{grad}}^{\delta}(\lambda_k)\|\|\lambda_k-\lambda_{k+1}\| + \left(\calL(\lambda_{k+1})-\calL(\lambda_{k})\right) \leq 0
\]
Then we have: 
\begin{align*}
    \|\lambda_{k+1}-\lambda_{k}\| & \leq \frac{1}{\beta}\left(\|\text{grad}(\lambda_k) - \tilde{\text{grad}}^{\delta}(\lambda_k)\| +  \sqrt{\|\text{grad}(\lambda_k) - \tilde{\text{grad}}^{\delta}(\lambda_k)\|^2 -2\beta\left(\calL(\lambda_{k+1})-\calL(\lambda_{k})\right) }\right)
\end{align*}
By denoting $B_k = \|\text{grad}(\lambda_k) - \tilde{\text{grad}}^{\delta}(\lambda_k)\|$ and $C_k = \calL(\lambda_{k+1})-\calL(\lambda_{k})$. Then we have:
\begin{align*}
    \|\lambda_{k+1}-\lambda_{k} \|^2 & \leq \frac{1}{\beta^2} \left(B_k^2 + B_k^2 -2\beta C_k + 2B_k \sqrt{B_k^2 - 2\beta C_k}\right)\\
    & \leq \frac{1}{\beta^2} \left(B_k^2 + B_k^2 -2\beta C_k + B_k^2 + B_k^2 - 2\beta C_k \right) \\
    & = \frac{4}{\beta^2} [\|\text{grad}(\lambda_k) - \tilde{\text{grad}}^{\delta}(\lambda_k) \|_2^2  - 2\beta \left(\calL(\lambda_{k+1})-\calL(\lambda_{k})\right)]
\end{align*}
Taking sum over $k$, we have:
\begin{align*}
    \sum_{k=1}^{+\infty} \|\lambda_{k+1}-\lambda_{k} \|^2 & \leq \frac{4}{\beta^2} \sum_{k=1}^{+\infty} \|\text{grad}(\lambda_k) - \tilde{\text{grad}}^{\delta}(\lambda_k) \|_2^2 -\frac{8}{\beta}(\lim_{k\to\infty} \calL(\lambda_{k+1})-\calL(\lambda_{1})) \\
    & \leq \frac{4}{\beta^2}\sum_{k}[(C+\kappa)^2\epsilon_k^{2} + \delta_{k}^2] -  \frac{8}{\beta}\left(\lim_{k\to\infty} \calL(\lambda_{k+1})-\calL(\lambda_{1})\right)< +\infty
\end{align*}
Since 1) the first term on the right side is finite, because the optimization tolerance is summable; 2) the second term is also finite, because the loss is assumed to be bounded. Then the upper bound is finite. In order to satisfy this condition, on the left side we should have:
\[
\lim_{k\to\infty} \lambda_{k+1}- \lambda_{k} = 0  
\]
By adopting the definition $\lambda_{k+1} = \text{Proj}(\lambda_{k}-\tilde{\text{grad}}^{\delta}(\lambda_k))$ and $\lim_{k\to\infty}\tilde{\text{grad}}^{\delta}(\lambda_k) = \text{grad}(\lambda_k)$ (Based on theorem 1, the limit of the optimization tolerance is zero), then we have:
\[
\lambda^{\star} = \text{proj}(\lambda^{\star}- \text{grad}(\lambda^{\star})) 
\]
Where $\lambda^{\star} = \lim_{k\to+\infty}\lambda_{k+1} = \lim_{k\to+\infty}\lambda_{k}$. Since the projection is on the bounded norm $L_{\text{norm}}$ and $\lambda^{\star}$ is within the bounded norm space, thus if $\lambda^{\star}- \text{grad}(\lambda^{\star})$ is within the bounded norm space, we have:
\[
\text{grad}(\lambda^{\star}) = 0
\]
Else if $\lambda^{\star}- \text{grad}(\lambda^{\star})$ is outside the bounded norm space, then according to the definition, the projection of $\lambda^{\star}- \text{grad}(\lambda^{\star})$ is surely on the \emph{boundary} of the $L_{\text{norm}}$ space, with $\|\text{proj}(\lambda^{\star}- \text{grad}(\lambda^{\star}))\|= L_{\text{norm}}$. However, we have assumed the $\lambda^{\star}$ is \emph{within} the bounded norm space with $\|\lambda^{\star}\|< L_{\text{norm}}$, which leads to the contradiction. Based on these discussions, we finally have:
\[
\text{grad}(\lambda^{\star}) = 0
\]
\end{proof}

\section{Possible extensions to non-binary protected features}
For the completeness, it is also possible to extend to binary protected features with distribution $D_1,\dots,D_N$. For example, the bi-level objective can be naturally formulated as
 
\[ \min_{\lambda} \sum_{n=1}^{N} \mathcal{L}_{n}(h^{\star}_{n},\lambda) + \sum_{m=1,n=1,n<m}^{n=N,m=N} \frac{\kappa_{n,m}}{2}\|h^{\star}_{n}-h^{\star}_{m}\|_2^2 \]

\[ \text{s.t.}\quad\quad\quad h^{\star}_{n} \in \text{argmin}_{h} \mathcal{L}_{n}(h,\lambda)
\]
 
Compared with the binary group, we introduce the pair-wise regularization ($\|h^{\star}_{n}-h^{\star}_{m}\|$) term to ensure the invariance between each pair of the sensitive attributes $(n,m)$. However, determining the coefficient $\kappa_{n,m}$ will become practically challenging, since the hyper-parameter space is much larger O($N^2$) than the binary case.
 
If the sensitive attribute is indeed a real value, a simple practical approach is to cluster the continuous attribute into several discrete groups, then conducting the pair-wise bi-level optimization. At the same time, there may be difficulty in measuring fairness. E.g, if the sensitive attribute is the ratio of people of certain demographic backgrounds, the corresponding sufficiency gap will be hard to estimate, since current metrics are defined on the discrete sensitive attribute.

\section{Additional Details and Results}
\subsection{Illustrative example of sufficiency gap}

\subsection{Correlation Analysis on the benchmark}
For the justification propose, we compute the Pearson correlation coefficient (ranging from $[-1,1]$) between the binary group index (or protected feature) $A$ and label $Y$. Intuitively, if $\D_{A=a}(Y=y)=\D_{A=a^{\prime}}(Y=y)$, $\forall a,a^{\prime},y$, the protected feature (or group index) $A$ is independent of label $Y$.

\begin{table}[h]
\caption{Pearson correlation coefficient between the group index and label}
\centering
\begin{tabular}{@{}cccc@{}}
\toprule
Toxic-Comment & CelebA & Law & NLSY \\ \midrule
0.30 & -0.35  & 0.18 &  -0.29 \\ \bottomrule
\end{tabular}
\end{table}    

The Pearson correlation coefficient clearly demonstrates the \emph{non-independence} between the group index and label. i.e, the label distributions among the sub-groups are different. The experimental results also validated this fact: \emph{the demographic parity based approach could not improve the sufficiency gap} due to the different label distributions.

\subsection{Additional Details}
\paragraph{Toxic Comments} We split the training, validation and testing set as $70\%$, $10\%$ and $20\%$. We adopt Adam optimizer with learning rate $10^{-3}$ and eps $10^{-3}$. The batch-size is set as 500 for each sub-group and we use sampling with replacement to run the explicit algorithm with maximum epoch 100. The fair coefficient is generally set as $\kappa=0.1\sim 0.001$. As for the inner-optimization step, the iteration number is 20 and the iteration in running conjugate gradient approach is 10. 

\paragraph{CelebA} The training/validation/test set are around $82$K, $18$K and $18$K. We also adopt the Adam optimizer with learning rate on $\lambda:10^{-5} \sim 10^{-4}$ and $h: 10^{-3}$. The batch-size is set as 64 for each sub-group and we iterate the whole dataset as one epoch. The maximum running epoch is set as $20$ and the iteration in running conjugate gradient approach is 10.

\paragraph{Law} We split the training, validation and testing set as $70\%$, $10\%$ and $20\%$. Then we adopt Adam optimizer with learning rate $10^{-3}$ and eps $10^{-3}$. The batch-size is set as 500 for each sub-group and we use sampling with replacement to run the implicit algorithm, with the maximum epoch 100. We adopt the MSE loss in the regression. The fair coefficient is generally set as $\kappa=0.1\sim 10^{-4}$. As for the inner-optimization, the iteration number is 20 and the iteration in running conjugate gradient is 10. In computing the sufficiency gap in the regression, we sample 33 points to compute the gap.
 
\paragraph{NLSY} We split the training, validation and testing set as $70\%$, $10\%$ and $20\%$. Then we adopt Adam optimizer with learning rate $10^{-3}$ and eps $10^{-3}$. The batch-size is set as 500 for each sub-group and we use sampling with replacement to run the implicit algorithm, with maximum epoch 100. We adopt the MSE loss in the regression. The fair coefficient is generally set as $\kappa=0.1\sim 10^{-4}$. As for the inner-optimization, the iteration number is 20 and the iteration in running conjugate gradient is 10. In computing the sufficiency gap, we sample 33 points to compute the sufficiency gap.

\subsection{Additional Empirical results}

\paragraph{Gradient evolution}
\begin{figure}[h]
    \centering
    \includegraphics[scale=0.5]{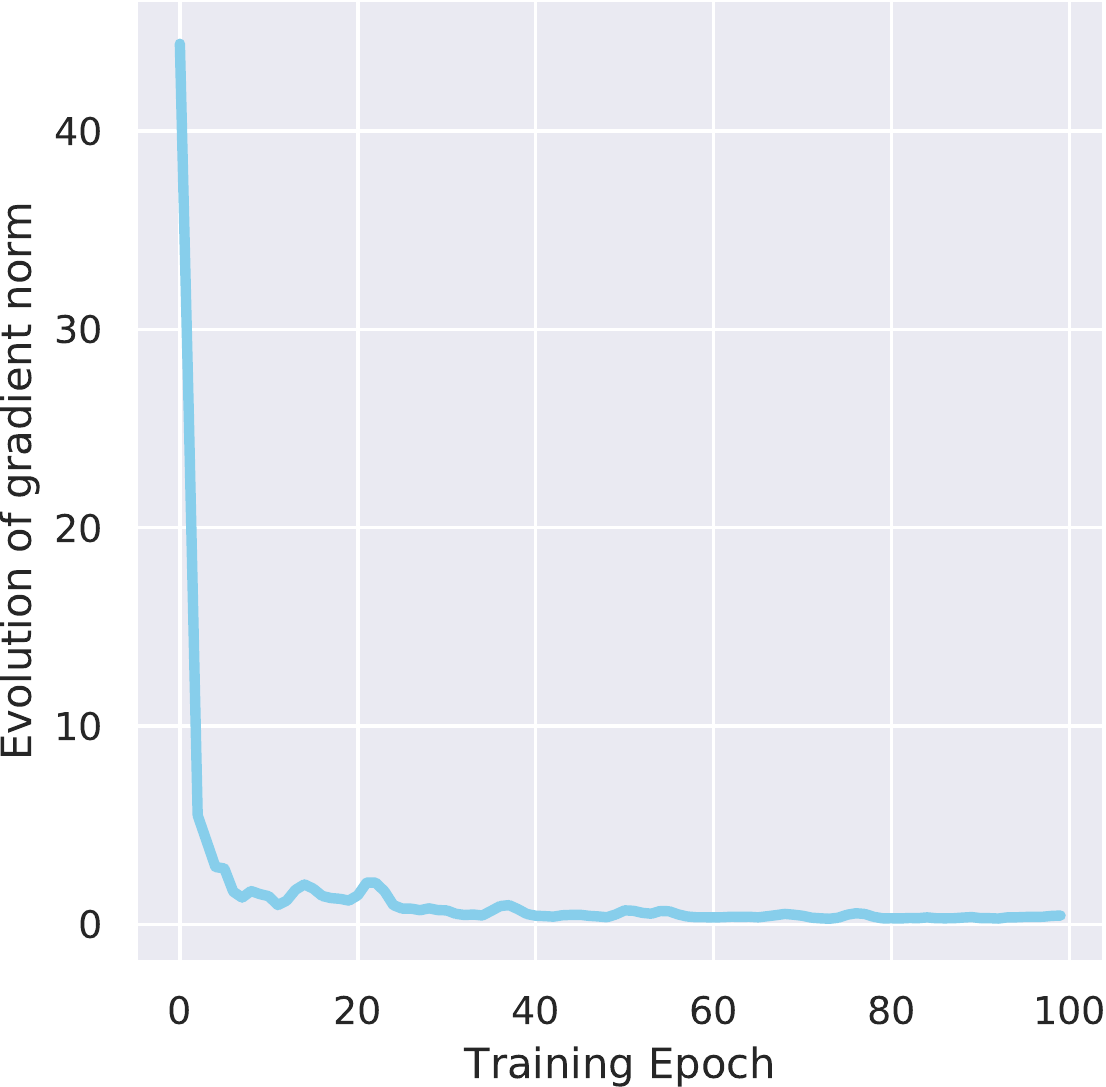}
    \caption{Gradient Norm evolution w.r.t. representation $\lambda$ in Toxic comments dataset. We visualize the norm of $\tilde{\text{grad}}^{\delta}(\lambda)$ at each training epoch, which suggests a convergence behavior and the gradient finally tends to zero.}
    \label{fig:grad_evolution}
\end{figure}
We also visualize the gradient norm of the representation $\lambda$ in the Toxic dataset, shown in Fig.~\ref{fig:grad_evolution}. The results verify the convergence behavior and the gradient norm finally tends to zero.

\subsection{Discussion with non-deep learning baselines}
In order to show the effectiveness of the proposed approach, we additionally compare the FAHT \citep{zhang2019faht}, a decision tree based fair classification approach. We evaluated the empirical performance on Toxic comments dataset. 
\begin{table}[h]
\caption{Comparison with Fairness Aware Decision Tree}
\centering
\begin{tabular}{@{}lcc@{}}
\toprule
Method & Accuracy ($\uparrow$)  & $\Delta\text{Suf}_C$ ($\downarrow$) \\ \midrule
FAHT     &  0.596	&  0.397 \\
Implicit & 0.760 & 0.051 \\ \bottomrule
\end{tabular}
\end{table}    

The implicit approach demonstrates the considerable better results, which may come from two aspects: (1) the Toxic task is a high-dimensional classification problem ($x\in\R^{748}$), where the deep learning based approach is more effective in handling the high-dim dataset. (2) The FAHT aims to realize the statistical parity (the independence rule), which is \emph{not compatible} with the sufficiency. According to the analysis of \citep{barocas-hardt-narayanan}, when the protected feature (A) and label (Y) are not independent (This has been justified by computing their Pearson Correlation coefficient), the sufficiency and independence cannot both hold.

\subsection{sufficiency Gap in regression}
We visualize the sufficiency gap of NLSY dataset.
\begin{figure}[t]
  \centering
  \subfigure[ERM]{\label{fig:erm-nlsy}\includegraphics[width=40mm]{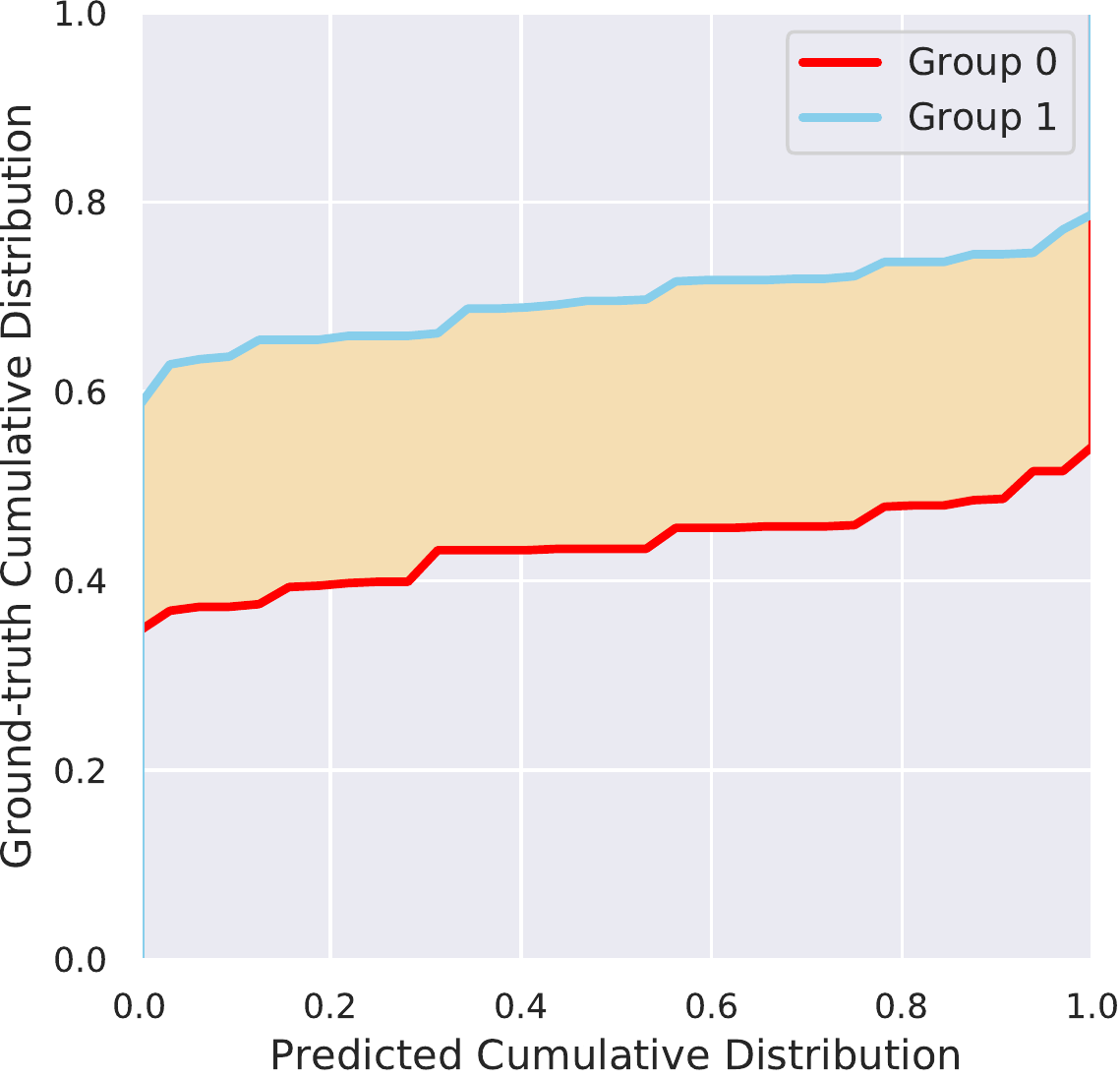}}
  \quad
  \subfigure[Fair Mix-up]{\label{fig:mix-up-nlsy}\includegraphics[width=40mm]{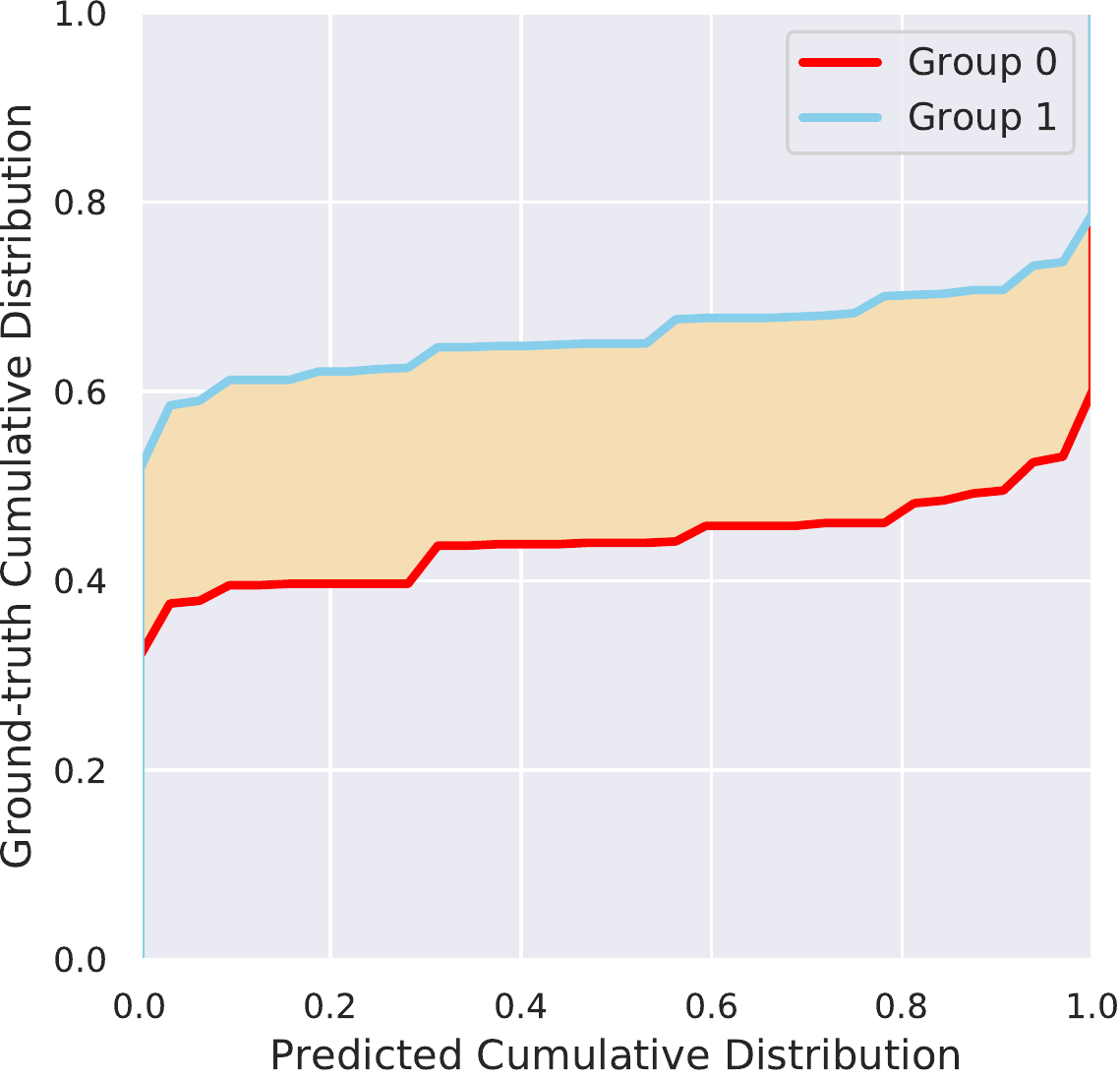}}
  \quad
  \subfigure[Implicit]{\label{fig:implict-nlsy}\includegraphics[width=40mm]{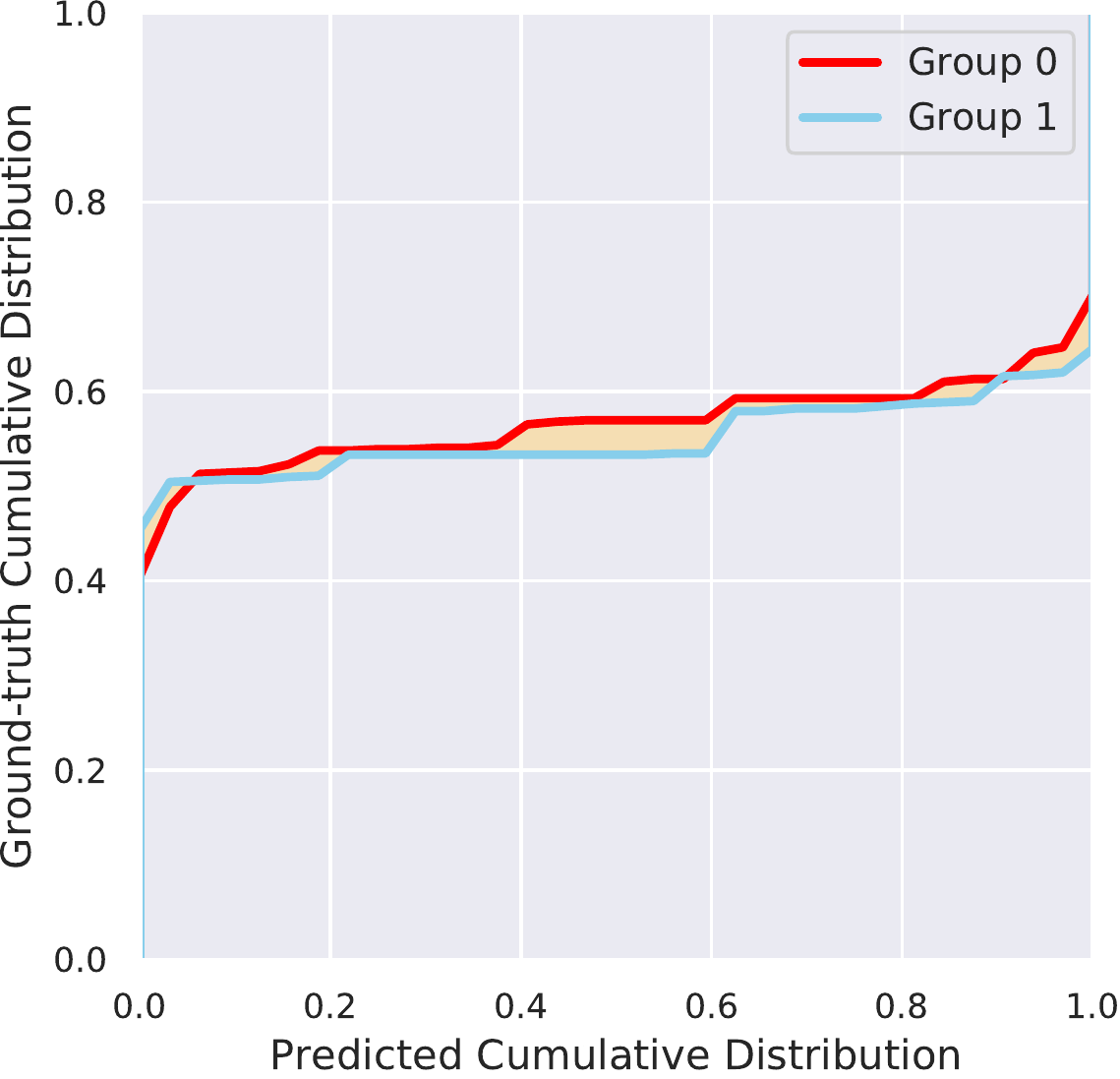}}
  \caption{Illustration of the \emph{sufficiency gap} in NLSY dataset. The ERM and mix-up suffer the high predictive sufficiency-gap, while the proposed implicit alignment can significantly mitigate the sufficiency gap. In contrast, the probability calibration is not improved. This results also verifies the inequivalence between the sufficiency gap and calibration gap \citep{pmlr-v97-liu19f}.}
  \label{fig:sufficient_gap_nlsy}
\end{figure}

\section{Complementary technical details}
We present complementary details that are related to the paper.
\subsection{Conjugate Gradient Method}
We present the Conjugate Gradient (CG) algorithm in Algo.~\ref{algo:cg_algo} through \verb+autograd+. In the conventional CG algorithm with objective $\frac{1}{2}x^T A X - bX$, we need to estimate $AX$ and compute its residual and update $X$. Since in our problem setting, the $A = \nabla^2_{h_0} \calL_0(h_0^{\epsilon},\lambda)$, then computing $AX$ can be realized through Hessian-vector product through \verb+autograd+, denoted as function $F$ in the paper. i.e., $\nabla^2_{h_0} \calL_0(h_0^{\epsilon},\lambda) X = F(x)$. 
\begin{algorithm}[t]
		\caption{Conjugate Gradient Method}
		\begin{algorithmic}[1]\label{algo:cg_algo} 
        \ENSURE Function $F$ that computes Hessian-vector product through \verb+autograd+, initial value $X_{0}$, bias vector $B$.
        \STATE Computing Residual: $r_0 = B - F(X_{0})$
        \STATE Set $p_0 = r_0$
        \FOR{inner\_iterations $k$}
        \STATE Computing $\alpha_k \leftarrow \frac{r_k^T r_k}{p_k^T F(p_k)}$
        \STATE $X_{k+1} \leftarrow X_{k} + \alpha_{k} p_{k} $
        \STATE $r_{k+1} \leftarrow r_{k} - \alpha_{k} F(p_{k})$
        \STATE If $r_{k+1}$ is sufficiency small, then stop.
        \STATE $\beta_{k} \leftarrow \frac{r_{k+1}^T r_{k+1}}{r_k^T r_k}$
        \STATE $p_{k+1} \leftarrow r_{k+1} + \beta_{k} p_{k} $
        \ENDFOR
        \STATE \textbf{Return:} $X_{k+1}$
        \end{algorithmic}
\end{algorithm}

Below we provided a simple PyTorch code for realizing the Hessian Vector product.

\begin{lstlisting}[language=Python, caption=Simple demo in computing Hessian vector product]
import torch
def hessain_vector_product(loss,model,vector):
    # loss: the defined loss
    # model: the model in computing the Hessian
    # vector: the required vector in computing Hessian-vector product
    partial_grad = torch.autograd.grad(loss, model_parameters(), create_graph=True)
    flat_grad = torch.cat([g.contiguous().view(-1) for g in partial_grad])
    h = torch.sum(flat_grad * vector_to_optimize)
    hvp = torch.autograd.grad(h, model.parameters())
    return hvp
\end{lstlisting}

\subsection{Calibration Gap in the regression}
Based on \citet{kuleshov2018accurate}, we first compute the predicted cumulative distribution ($\hat{Y}_0$) of at point $t$: $D_0(\hat{Y}_0\leq t)=\alpha$, then we compute the corresponding ground truth cumulative distribution ($Y_0$)  at point $t$. By changing $t$, we obtain several points on function $\D_0(Y\leq t|\hat{Y}_0\leq t)=\beta$. Then the regression is probabilistic calibrated when $\alpha\equiv\beta$. From this perspective, the zero calibration gap can guarantee a zero sufficiency gap. But the inverse is not necessarily true, as our experimental results suggest, a small sufficiency gap can lead to either small or large calibration gap. 
Thus it can be quite promising to explore their inherent relations and trade-off in the fair regression.





\end{document}